\documentclass{article}

\PassOptionsToPackage{numbers, compress}{natbib}
 \usepackage[main, final]{neurips_2026}
\usepackage[utf8]{inputenc} % allow utf-8 input
\usepackage[T1]{fontenc}    % use 8-bit T1 fonts
\usepackage{hyperref}       % hyperlinks
\usepackage{url}            % simple URL typesetting
\usepackage{booktabs}       % professional-quality tables
\usepackage{amsfonts}       % blackboard math symbols
\usepackage{nicefrac}       % compact symbols for 1/2, etc.
\usepackage{microtype}      % microtypography
\usepackage{xcolor}         % colors
\usepackage{amsmath}
\usepackage{amssymb}
\usepackage{mathtools}
\usepackage{amsthm}
\usepackage{wrapfig}
\usepackage{xcolor}
\usepackage{enumitem}
\usepackage{graphicx}
\usepackage{subcaption}
\usepackage{multirow}
\usepackage{gensymb}
\usepackage{graphicx}
\definecolor{emgpurple}{HTML}{8E6FB8}
\definecolor{emgred}{HTML}{E0625B}
\definecolor{videogreen}{HTML}{3BA86B}

\title{\textsc{MotorSense}: A Video-EMG Dataset of Motoric Representations for Action Understanding}

\title{MotorSense: A Video-EMG Dataset of Motoric Representations for Action Understanding}

\author{%
  \textbf{Eadom Dessalene}$^{1}$ \quad
  \textbf{Michael Maynord}$^{1}$ \quad
  \textbf{Amir-Hossein Shahidzadeh}$^{1}$ \\
  \textbf{Botao He}$^{1}$ \quad
  \textbf{Yasamin Tabatabaee}$^{2}$ \quad
  \textbf{Ruoshi Liu}$^{1}$ \\
  \textbf{Ruohan Gao}$^{1}$ \quad
  \textbf{Paul Pu Liang}$^{3}$ \quad
  \textbf{Yiannis Aloimonos}$^{1}$ \\[0.6em]
  $^{1}$University of Maryland, College Park \quad
  $^{2}$Carnegie Mellon University \\
  $^{3}$Massachusetts Institute of Technology
}

\begin{document}

\maketitle
\vspace{-2.4em}
\begin{center}
\small
\href{https://www.cs.umd.edu/~edessale/motorsense/}{\texttt{https://www.cs.umd.edu/\textasciitilde edessale/motorsense/}}
\end{center}
% EMG provides a dense hand signal free of both occlusion and labeler bias.

\begin{abstract}
Computer vision models learn hand action only from its visual consequences, without access to the underlying embodied motor signals. We address this absence with \textsc{MotorSense} - the first large-scale paired egocentric video and surface electromyography (EMG) dataset. \textsc{MotorSense} pairs bimanual EMG bands with egocentric video and yields synchronized video, bimanual EMG and IMU, and audio across $100$ hours of unscripted manipulation. EMG offers a powerful scalable alternative to manual annotation: it captures finger movement, hand configuration, and the forces exerted between hand and environment as a dense signal free of occlusion and labeler bias, whereas manual annotation is costly, subjective, and unreliable under occlusion. We use \textsc{MotorSense} to learn \textit{sensorimotor alignment}, producing an organization of action that is shared across the visual and motoric spaces, through cross-modal pre-training between video and EMG. The resultant vision models are then transferred to downstream tasks that benefit from \textit{sensorimotor alignment}: in this case, action recognition, 3D hand reconstruction, and 3D contact reconstruction. We further demonstrate a generative application in which EMG motor signals are used to condition a video diffusion model, yielding videos of fine-grained hand manipulation. Across four action-centric computer vision tasks and eight popular video datasets, we demonstrate that \textit{sensorimotor alignment} provides a stronger foundation for visual representation learning than conventional self-supervised learning methods over video, outperforming every baseline across all tasks and datasets evaluated. 
\end{abstract}

\section{Introduction}

When humans develop motor and perceptual capabilities, they rely not only on perceiving their surroundings and the visual effects of their actions, but more crucially on the internal motor experiences that accompany these percepts. Yet computer vision models are trained almost exclusively on the visual modality, with no direct access to embodied modalities such as muscle activity. Closing this gap requires incorporating such signals: learning from video alone leaves the underlying action structure implicit in a high-dimensional stream of pixels, whereas incorporating embodied signals transforms this hidden structure into an intrinsically lower-dimensional learnable signal. Among embodied signals, those tied to the hands are especially important, as it is through the hands that humans effect change in the physical world. 

\begin{figure}[t!]
  \centering
  \includegraphics[width=1\textwidth]{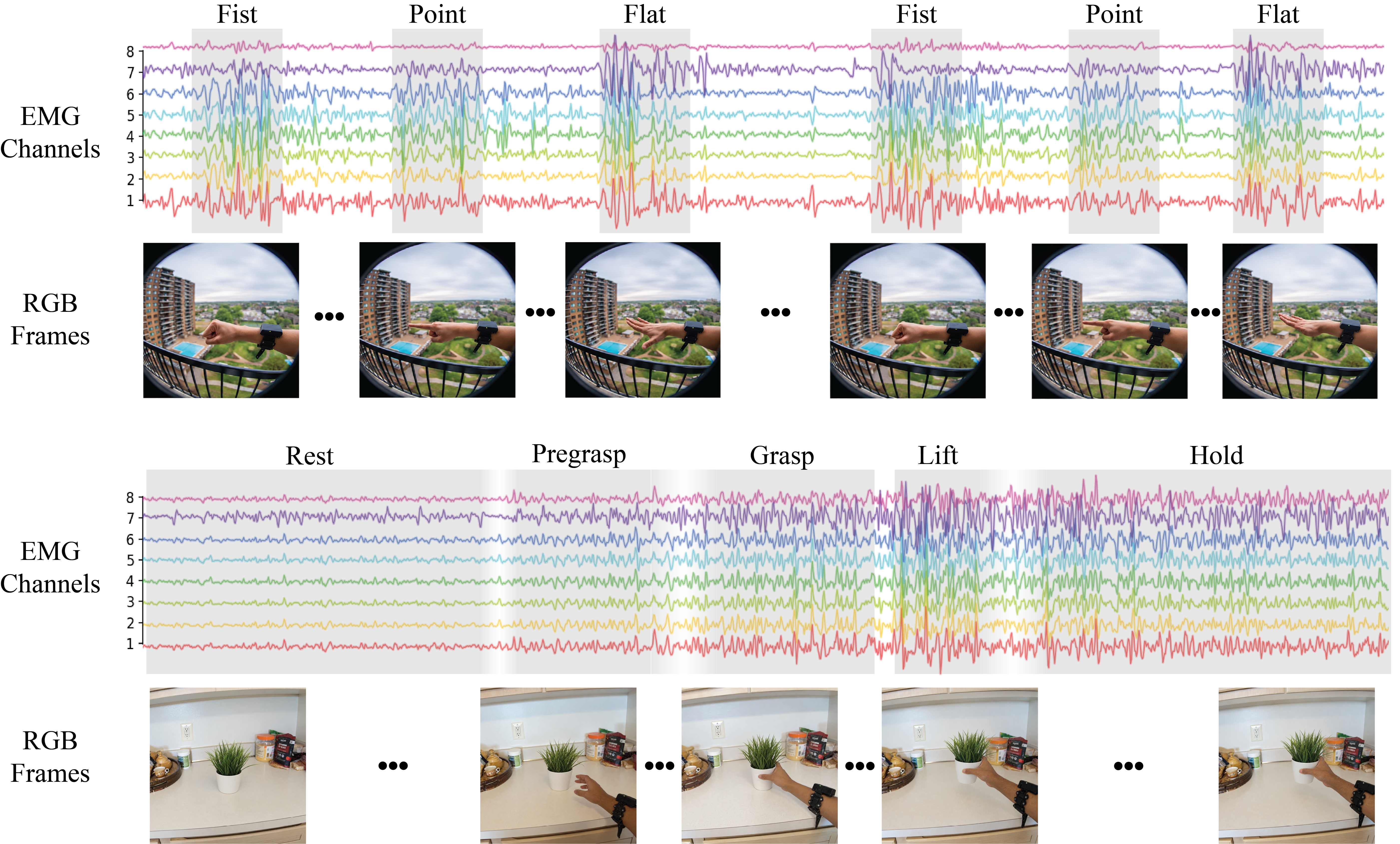}
  \caption{\textbf{EMG captures both categorical and continuous structure of hand actions.} \textbf{(Top)} For three discrete hand configurations repeated twice---\textit{fist}, \textit{point}, and \textit{flat}---each pose produces a distinct and reproducible EMG signature: \textit{fist} is marked by a suppressed \textcolor{violet}{\textbf{purple}} channel ($\#7$) with respect to \textit{flat} and a pronounced \textcolor{emgred}{\textbf{red}} channel ($\#1$) with respect to point; \textit{point} exhibits lower overall amplitude, with less \textcolor{violet}{\textbf{purple}} than \textit{flat} and less \textcolor{emgred}{\textbf{red}} than \textit{fist}; \textit{flat} is distinguished by pronounced \textcolor{violet}{\textbf{purple}} and \textcolor{emgred}{\textbf{red}} activation. Sharp grey rectangles reflect the boundaries between hand gestures and rest states. \textbf{(Bottom)} A pick-and-lift sequence combines gradual and abrupt transitions: the hand moves smoothly from \textit{rest} through \textit{pregrasp} into \textit{grasp}, changes sharply as force is applied at \textit{lift}, then settles---non-instantaneously---into a steady \textit{hold}. The EMG mirrors this structure, with channels quiet at rest, building subtly through pregrasp and grasp, peaking sharply as the lift begins, and stabilizing during hold. }
  \label{fig:sequence_examples}
\end{figure}

To make hand-centric embodied signals available to computer vision we introduce \textsc{MotorSense}, a dataset pairing egocentric video with direct measurements of hand motor activity collected via surface electromyography (EMG) — a modality that records the electrical activity of muscles through the skin. See Figure \ref{fig:sequence_examples} for example sequences. \textsc{MotorSense} captures a dense signal reflecting how the entirety of the hand engages with the environment: the forces applied, finger movement, and hand configuration throughout each interaction, fine-grained information difficult to observe directly from vision due in part to occlusion. \textsc{MotorSense} goes beyond force datasets such as FEEL \cite{dessalene2026feel}, TouchScale \cite{li2026touchscale}, and OpenTouch \cite{song2025opentouch} with a richer and denser embodied signal, gathered without the use of hand-mounted sensors which visually and physically obstruct interaction. The direct measurement in \textsc{MotorSense} stands in contrast to labels annotated over video datasets (e.g. EPIC-Kitchens, Ego4D), which 1) require manual annotation effort, 2) require pre-defined ontology design, and 3) suffer from annotation error and imprecision.

\textsc{MotorSense} captures hand motor activity using EMG devices consisting of 8 equally spaced electrodes around the lower forearm (see Figure \ref{fig:emg_hardware}). EMG directly senses the electrical activity produced by muscles through the skin, where different muscles produce distinct activation profiles across the electrode array. EMG signals encode both hand forces \cite{duque1995evaluation} and fine-grained, dynamic finger articulation \cite{salter2024emg2pose}. In addition to EMG, we record IMU streams from the wrist device capturing translational and rotational acceleration of the hand in space, and synchronize these signals with egocentric video. 

% Together, these streams form a time-synchronized multimodal record of natural manipulation — one in which the structure of hand action lies latent within the continuous motor signal.

Our primary use of \textsc{MotorSense} is representation learning through \textit{sensorimotor alignment} between video and EMG data. We induce this alignment with a CLIP-style contrastive objective that jointly trains a video encoder and an EMG encoder in a shared embedding space, pulling together each video clip and the muscle activity that produced it and pushing apart mismatched pairs; EMG thus acts as a privileged supervisory signal that shapes the video representation during pre-training but is discarded at inference, so the aligned video backbone can be transferred to purely visual downstream tasks. We transfer this backbone to three tasks that emphasize hand--object manipulation: 1) recognizing actions in video; 2) 3D hand reconstruction from images; 3) 3D contact reconstruction from images.

Beyond representation learning, \textsc{MotorSense} also supports a generative application: conditioning an image-to-video (I2V) diffusion model on EMG. This addresses a key limitation of existing control modalities, which cannot specify fine-grained, physically grounded hand actions --- language captures high-level activity but is underspecified, keypoint trajectories require annotation and are force-agnostic, and neither captures the fine-grained forces of manipulation. EMG supplies exactly this signal, making generative models more useful for VR/AR, where EMG wearables are already emerging as natural control interfaces \cite{zheng2022surface}.

The primary contributions of this work are as follows:
\begin{itemize}[leftmargin=1.2em]

\item \textbf{Release of the \textsc{MotorSense} dataset}: We release the first large-scale paired egocentric video and surface electromyography (EMG) dataset, capturing the muscle activations that cause physical action rather than only their visual effects. \textsc{MotorSense} consists of $100$ hours of natural, unscripted hand–object manipulation, recorded with non-obstructive wrist-worn EMG bands paired with egocentric glasses.

\item \textbf{Sensorimotor alignment for action-centric computer vision:} We demonstrate that aligning a video encoder with bimanual EMG data through a CLIP-style contrastive objective is a powerful self-supervised pre-training approach, yielding large and consistent gains over a variety of visual-only pre-training methods.

\item \textbf{Motor-code conditioning for fine-grained control of video diffusion:} We learn motor codes, a discrete VQ-VAE representation of EMG discovered without manual labels, and use them as conditioning inputs to an image-to-video diffusion model. This allows for video generation responsive to fine-grained hand activity, and improves generation fidelity for hand–object interaction.

\end{itemize}

\section{Related Work}
\label{sec:related_work}

\textbf{Current Applications of EMG Sensing.} Surface electromyography (EMG) measures muscle activation through skin-mounted sensors~\cite{merletti2004electromyography, farina2014extraction}, with forearm placement particularly informative for hand actions since finger flexion, extension, wrist articulation, and grasp force produce distinct activation patterns~\cite{duque1995evaluation, de1997use}. EMG thus provides direct access to motor activity that is often hidden or ambiguous in video. Prior work has applied EMG to gesture recognition~\cite{atzori2014ninapro, geng2016gesture,xi2026egoemg}, grasp classification and reach-to-grasp decoding~\cite{batzianoulis2017grasp}, finger and grasp force estimation~\cite{castellini2012fingerforce,he2026forceband}, and lightweight VR/AR input via wrist-worn sensors~\cite{salter2024emg2pose, sivakumar2024emg2qwerty, kaifosh2025generic}. The only prior vision dataset pairing EMG with video in capturing object manipulation, ActionSense~\cite{delpreto2022actionsense}, is limited to a few action types in a constrained lab setting, uses post-processed armband EMG ill-suited to fine-grained manipulation, and never integrates EMG into video learning. We close this gap by pairing egocentric video with wrist-worn EMG, using it as a cross-modal supervisory signal to pre-train video models through contrastive alignment, and using EMG-derived motor codes as a control signal for generative video.

\textbf{Multimodal Signals for Action Understanding in Video.} We organize prior work along the three downstream applications we target with \textsc{MotorSense}: action recognition, 3D hand-object reconstruction, and video diffusion. For action understanding, non-visual signals shown to carry transferable structure include language~\cite{pramanick2023egovlpv2, zhao2023learning, wang2024internvideo2}, audio~\cite{morgado2021audio, alwassel2018action}, head- and body-mounted IMU~\cite{moon2023imu2clip, zhang2024masked, chen2025comodo}, 3D body and hand pose~\cite{thoker2019cross, shamil2024utility}, and contact force~\cite{dessalene2026feel, song2025opentouch,zhou2026touchanything}; we instead leverage EMG to capture the muscle activity underlying manipulation directly. For 3D hand-object reconstruction, prior work is almost exclusively visual---monocular RGB~\cite{pavlakos2024reconstructing, potamias2025wilor, fan2024hold} or depth/RGB-D~\cite{zhang2021single, chen2023tracking, garcia2018first, yuan2018rgb}---with rare audio-based exceptions generating contact sounds from trajectories~\cite{dou2025hearing} or recovering pose and per-vertex contact from wearable microphones~\cite{mao2025visuo}; EMG offers an automatic, occlusion-free measurement of finger configuration and contact. For video diffusion, existing conditioning signals---language~\cite{yang2024cogvideox, kong2024hunyuanvideo, wan2025wan}, keypoints~\cite{zhang2024mimicmotion, li2024dispose, gan2025humandit}, and audio~\cite{zhu2023taming}---each fall short for contact-rich manipulation: language is semantically underspecified, keypoints are noisy near contact and force-agnostic, and audio reflects scene acoustics rather than motor activity. EMG supplies the fine-grained, force-aware control these modalities lack.

\textbf{Understanding Hands in Video.} Large-scale egocentric datasets~\cite{damen2018scaling, grauman2022ego4d, grauman2024egoexo4d} and contact annotations~\cite{darkhalil2022epic} provide hand-object supervision but rely on coarse, costly manual labels. Multi-view capture rigs~\cite{hampali2021ho, chao2021dexycb, kwon2021h2o, liu2022hoi4d, fan2023arctic, ohkawa2023assemblyhands, banerjee2025hot3d} recover high-fidelity pose but leave contact forces unobserved and depend on laboratory infrastructure that limits in-the-wild collection. Sensorized gloves~\cite{dessalene2026feel, song2025opentouch, delpreto2022actionsense} record kinematics, contacts, and applied force directly, but obstruct natural hand use. Consumer-grade surface EMG and FMG wristbands~\cite{castellini2009surface, sivakumar2024emg2qwerty, salter2024emg2pose,kolev2026beyond} sidestep this trade-off, capturing finger movement, hand configuration, and exerted forces in unconstrained settings at a scale lab setups cannot match.

\section{Methods}
\label{sec:methods}
\begin{figure}[t!]
  \centering
\includegraphics[width=1\textwidth]{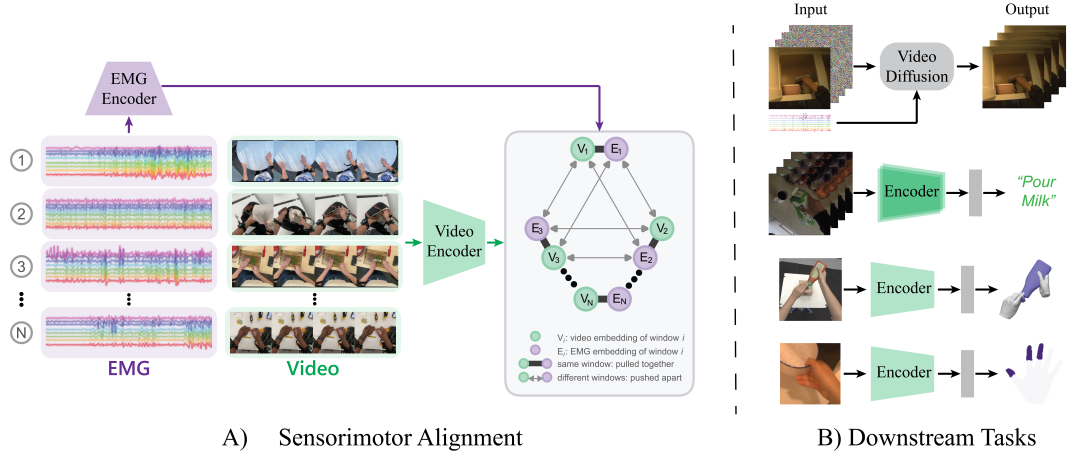}
  \caption{Overview of \textsc{MotorSense}. \textbf{(A) Sensorimotor Alignment.} Windows $1,\dots,N$ of bimanual EMG (left) and egocentric video (right) are embedded by an EMG encoder and a video encoder and aligned with a symmetric contrastive objective (Section~\ref{sec:emg_pretraining}): the video and EMG embeddings of the same window ($V_i$, $E_i$) are pulled together, while embeddings from different windows are pushed apart. \textbf{(B) Downstream Tasks.} The learned representations are used for downstream tasks: 1) video diffusion consisting of a single reference frame as input along with EMG control signals, 2) action recognition, 3) 3D hand-object Reconstruction, 4) 3D hand contact prediction.}

  \label{fig:task_description}
\end{figure}

This section presents the components that together define \textsc{MotorSense} and the learning framework built on top of it. We first describe the data collection setup, which pairs bimanual surface EMG and IMU with synchronized egocentric
video from glasses
(Section~\ref{sec:dataset_collection}). We then present our primary use of the dataset,
\emph{sensorimotor alignment}, or the alignment of visual embeddings and EMG motor embeddings, resulting in unimodal vision networks that produce representations that are reflective of the fine-grained motor structure of manipulation (Section~\ref{sec:emg_pretraining}).
Separately, for our generative application, we introduce \emph{motor codes} --- a discrete
mid-level representation of EMG learned directly from the raw signal via a VQ-VAE, with no
manual action labels (see Appendix for details) --- and use them as conditioning inputs for an image-to-video
diffusion model, providing a fine-grained, muscle-derived control signal that existing
modalities such as language and keypoints cannot supply (Section~\ref{sec:diffusion_methods}).
For a depiction of the contrastive pre-training pipeline and the downstream tasks, see
Figure \ref{fig:task_description}.

\begin{wrapfigure}{r}{0.6\textwidth}
  \centering
  \vspace{-\baselineskip}
  \includegraphics[width=0.6\textwidth]{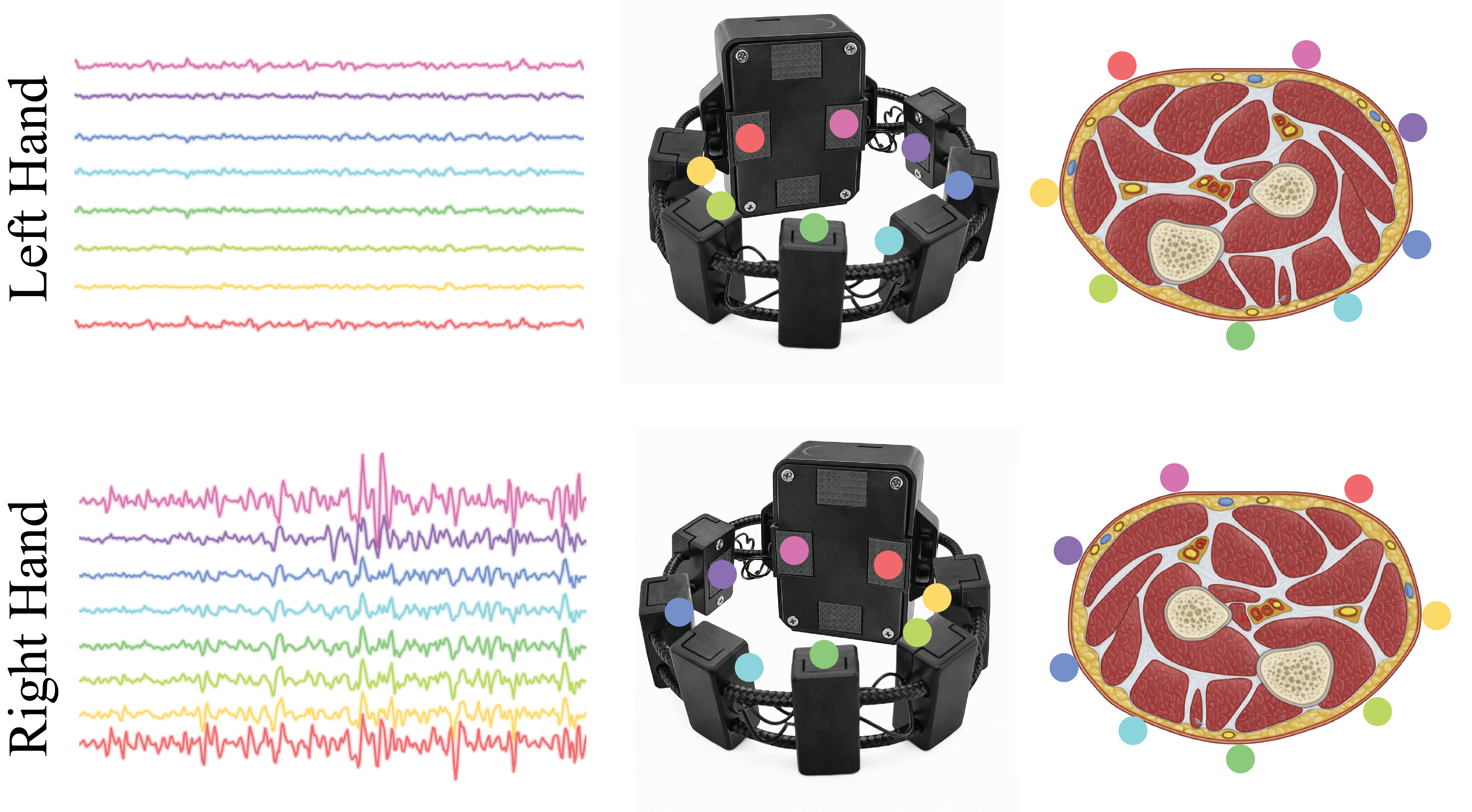}
  \caption{The bimanual EMG sensing setup of \textsc{MotorSense}. \textbf{Middle}: the MindRove Armband used for data collection, worn on the left and right wrists (top and bottom row, respectively), with colored dots marking the electrode sensors equidistantly arranged around the band. \textbf{Right}: cross-sectional anatomy of the lower forearm with each electrode overlaid. \textbf{Left}: example 8-channel EMG recordings from the left and right wristbands gathered during a manipulation task primarily involving the right hand; each colored trace corresponds to one electrode. }
  \label{fig:emg_hardware}
\end{wrapfigure}

\subsection{Dataset Collection}
\label{sec:dataset_collection}
\textsc{MotorSense} pairs bimanual surface EMG with egocentric video in a wearable, non-obstructive setup that captures natural hand--object manipulation. We collect data in kitchen environments, which provide a broad range of object interaction and action types. For details on dataset statistics and dataset collection procedures pertaining to the data recorded by Aria glasses and the data processing of both vision and EMG streams, see Section \ref{appendix:dataset_collection} within the Appendix.

\textbf{EMG} To capture forearm muscle activity during manipulation, we use the MindRove Armband. Two wireless surface EMG devices are worn just slightly above the wrists. See Figure \ref{fig:emg_hardware} for a picture of the EMG. The armband provides 8 EMG channels arranged equidistantly in a circular configuration around the forearm, with two additional bias and reference channels, using dry stainless-steel electrodes that eliminate the need for conductive gel or skin preparation. Signals are sampled at $500$ Hz with $24$-bit resolution and streamed over Wi-Fi to the recording workstation. The device also includes an integrated 6-axis IMU (3-axis accelerometer and 3-axis gyroscope) sampled at $50$ Hz, which we retain as an auxiliary wrist motion signal aligned to the EMG stream. The circumferential electrode layout provides broad coverage of both the flexor and extensor compartments, capturing activity from the muscle groups most relevant to grasping, finger flexion/extension, and wrist articulation during object manipulation.

\subsection{Sensorimotor Alignment through CLIP-Style Pretraining}
\label{sec:emg_pretraining}

% EMG provides a dense, occlusion-free measurement of the motor activity underlying hand action. A natural question is whether a visual encoder can be shaped by this signal, so that its representations become sensitive to the fine-grained motor structure of manipulation, and whether the learned representations transfer to action-centric vision tasks. We pursue this through \emph{sensorimotor alignment}: we jointly train a visual encoder and an EMG encoder to embed video and EMG windows into a shared space, pulling together the embeddings of video--EMG pairs and pushing apart mismatched pairs. As in CLIP-style alignment, EMG acts as a privileged supervisory signal available only at pre-training time: after alignment, the visual encoder is transferred to downstream tasks on its own, with no EMG required at inference.

% Each training sample is a pair consisting of a short egocentric RGB clip and the bimanual EMG + IMU window. Positives are temporally aligned video--EMG pairs drawn from the same recording; all other pairs within the batch serve as negatives.

EMG provides a dense, occlusion-free measurement of the motor activity underlying hand action. A natural question is whether a visual encoder can be shaped by this signal to become sensitive to the fine-grained motor structure of manipulation, and whether the learned representations transfer to action-centric vision tasks. We pursue this through \emph{sensorimotor alignment}, jointly training visual and EMG encoders to embed windows of egocentric RGB video and bimanual EMG + IMU into a shared representation space. Following CLIP-style cross-modal pre-training, EMG serves as a privileged supervisory signal available only during pre-training. After alignment, the visual encoder is transferred independently to downstream action-centric vision tasks, without requiring EMG at inference.

\textbf{Vision Tower} A vision backbone encodes the clip, over which we perform spatial pooling (for video networks we do \textit{not} perform temporal pooling, which we find to be a crucial design choice in Section \ref{sec:imu_ablation}). See Appendix for details.

\textbf{EMG Tower} The EMG tower is a temporal convolutional--recurrent network. The early stages consist of strided 1D convolutions, followed by residual depthwise-convolution blocks and an LSTM model for encoding of temporal structure. The EMG input is padded with $1$ second of context around the video window. IMU is encoded by a second tower of the same design and aligned to video with its own contrastive term. See Appendix for details.

\textbf{Objective and Loss} Because EMG signals exhibit recording-specific biases, we construct each training batch by sampling windows from the same recording. Without this constraint, the visual encoder can minimize the contrastive loss by identifying recording sessions rather than learning action-relevant features (see Appendix for details).  As in CLIP, visual and EMG embeddings are $L_2$-normalized and compared by cosine similarity, scaled by a learnable temperature $\tau$ (a learned logit scale). We minimize the symmetric InfoNCE (CLIP-style) loss
\begin{equation}
  \mathcal{L} = \frac{1}{2}\big(\mathcal{L}^{\mathrm{EMG}} + \mathcal{L}^{\mathrm{IMU}}\big), \qquad
  \mathcal{L}^{m} = \frac{1}{2}\big(\mathcal{L}^{m}_{v \rightarrow m} + \mathcal{L}^{m}_{m \rightarrow v}\big),
  \label{eq:loss}
\end{equation}
where $\mathcal{L}_{v\rightarrow e}$ and $\mathcal{L}_{e\rightarrow v}$ are cross-entropy losses over the vision-to-EMG and EMG-to-vision similarity matrices respectively, each taking the matched pair on the diagonal as the target.

\textbf{Transfer} After pre-training, we discard the EMG tower, retain the vision backbone, attach task-specific layers (a classification head for action recognition, and regression heads for 3D hand reconstruction and 3D contact reconstruction), and fine-tune on the downstream tasks (Section~\ref{sec:downstream_eval_exp}).

\subsection{Video Diffusion Conditioned on Motor Codes}
\label{sec:diffusion_methods}

For our generative application, motor codes serve as conditioning inputs to a generative video model. We first learn the motor codes directly from raw EMG + IMU data. Given a multichannel EMG + IMU window, we train an encoder--quantizer--decoder model that follows a VQ-VAE design and maps the window to a sequence of discrete tokens, each indexing into a shared learned codebook; this token sequence is the motor-code representation of the window. The model is trained with a standard VQ-VAE objective combining a reconstruction loss on the input window with a latent commitment penalty that pulls the encoder's continuous outputs toward their assigned codebook entries. For full implementation details, including model architecture, see Section~\ref{sec:appendix_code_discovery} of the Appendix.

% \begin{wrapfigure}{r}{0.4\textwidth}
%   \centering
%   \vspace{-\baselineskip}
%   \includegraphics[width=0.4\textwidth]{VQVAE.png}
%   \caption{We follow the VQ-VAE \cite{van2017neural} framework, where a multichannel EMG window is 1) fed to an encoder, 2) quantized against a jointly learned codebook to produce a sequence of discrete motor codes, the embeddings of which are 3) fed to the decoder for reconstruction of the EMG window. Training is self-supervised, and IMU is included but not shown. {\color{red} fix arrows from video encoder, fix spacing differences}}
%   \label{fig:motor_codes}
% \end{wrapfigure}

We consider an image-to-video (I2V) task: given a single reference frame and a sequence of motor codes, predict the subsequent video clip. Rather than fully fine-tuning a large video diffusion backbone on paired video--motor-code data, which would be prohibitively expensive, we leverage Video2LoRA \cite{wu2026video2lora}, a lightweight approach for controlling video diffusion networks, repurposed to make use of our motor codes as a novel control modality.

\textbf{Problem Setup} Each training sample is a tuple of a reference frame (the first frame of a clip) and a pair of bimanual motor-code sequences extracted from the VQ-VAE spanning the clip duration.

\textbf{HyperNetwork-Predicted Lightweight LoRA} We follow the LightLoRA design of Video2LoRA~\citep{wu2026video2lora}: the encoded motor-code sequence is fed to a HyperNetwork, which outputs, in a single forward pass, a bank of sample-specific prediction matrices --- one pair per controlled linear layer in the diffusion backbone. We refer the reader to \cite{wu2026video2lora} for the full details, including the exact factorization and the iterative refinement procedure used by the HyperNetwork. The only trainable components are a learned motor-code embedding network, a HyperNetwork that maps encoded motor codes to adapter weights, and the low-rank adapters inserted into the video diffusion backbone.

\begin{figure*}[t]
  \centering
  \includegraphics[width=1.0\textwidth]{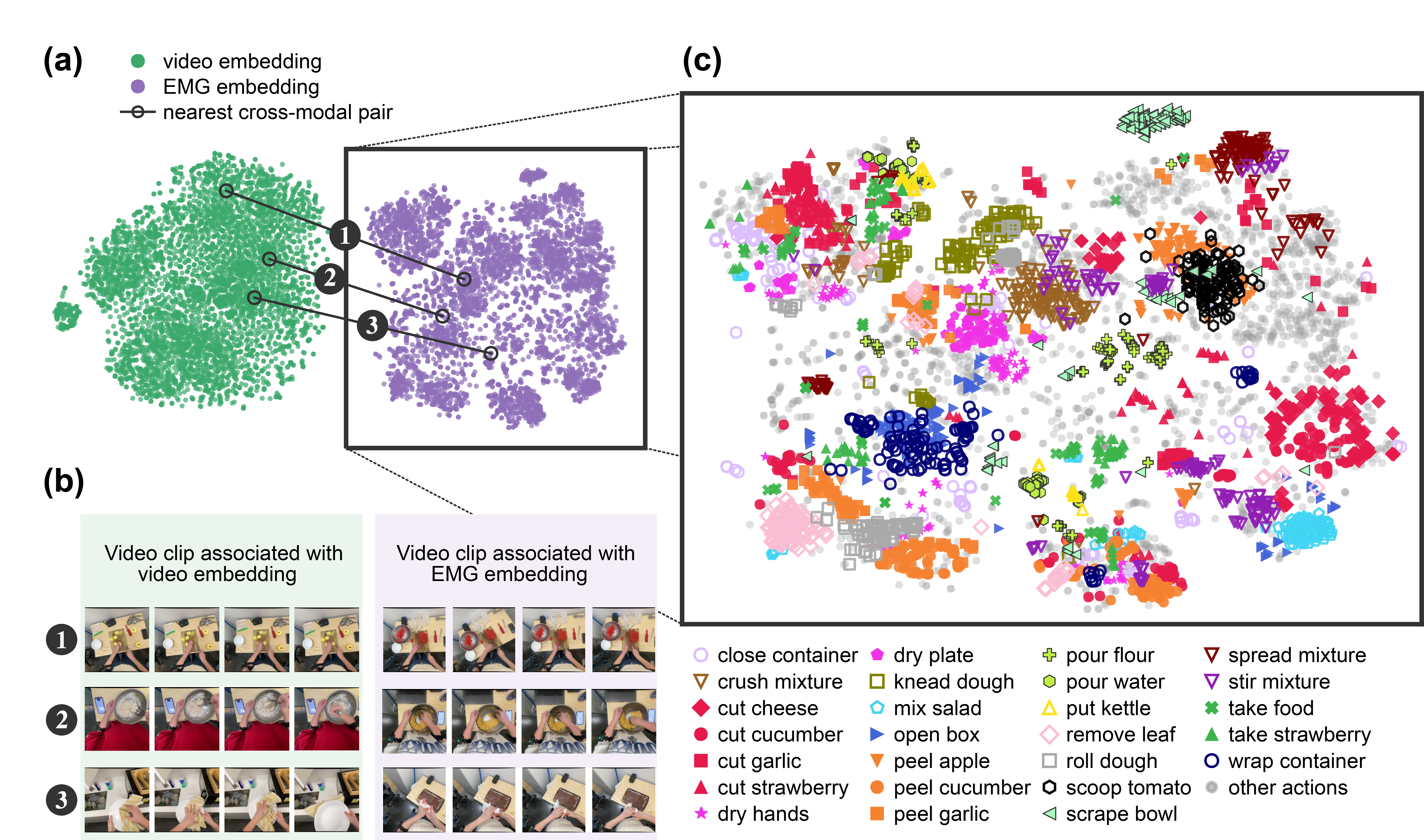}
  
\caption{\textbf{Self-supervised sensorimotor alignment organizes video and EMG by hand action.}
\textbf{(a)} t-SNE plot of the joint embedding space for video ({\color{green!50!black}green}) and EMG ({\color{emgpurple}purple}) derived from the two towers of our self-supervised contrastive model (Section~\ref{sec:emg_pretraining}). A "modality gap" is observed ~\citep{liang2022mind} between video and EMG clusters. \textbf{(b)} The clips associated with each pair: the video query (left) and the clip whose EMG was retrieved (right). Each retrieved clip performs the query's hand action---cutting, mixing in a bowl, drying---with a different object, scene, and recording.
\textbf{(c)} The EMG embeddings of (a), coloured by verb with marker shape denoting the object. Despite the network never being trained over these labels, windows of the same action form compact regions (e.g.\ \textit{mix salad}, \textit{scrape bowl}) and verbs group together (the \textit{cut} family in red). Best viewed zoomed-in and in colour.}

\label{fig:sensorimotor_tsne}
\end{figure*}

\section{Experiments}
\label{sec:experiments}
Here we present experiments performed over \textsc{MotorSense}. We first evaluate sensorimotor alignment learning of image and video backbones, and transfer the resulting representations to three downstream task families (Section~\ref{sec:downstream_eval_exp}) chosen for their emphasis on hand--object manipulation: action recognition on Something-Something v2 \cite{goyal2017something},
  EPIC-Kitchens-100 \cite{damen2018scaling}, Ego-Exo4D \cite{grauman2024egoexo4d}, and \textsc{MotorSense}; 3D hand reconstruction on ARCTIC \cite{fan2023arctic}, H2O \cite{kwon2021h2o}, and HOT3D
  \cite{banerjee2025hot3d}; and 3D hand contact on \textsc{MotorSense}. Finally, we evaluate motor codes as a conditioning signal for a video diffusion model (Section~\ref{sec:exp_video_diffusion}), comparing performance against an unconditioned image-to-video backbone and a language conditioning baseline derived procedurally from the raw EMG and IMU streams. We also analyze the sensorimotor
  embedding space (Section~\ref{sec:exp_sensorimotor_analysis}).

  \begin{table*}[t]
  \centering
  \setlength{\tabcolsep}{6pt}
  \renewcommand{\arraystretch}{1.08}
  \small
  \resizebox{\textwidth}{!}{%
  \begin{tabular}{c l ccc c c c}
  \toprule
   & & \multicolumn{3}{c}{\textbf{EPIC-Kitchens-100}} & \textbf{SSv2} & \textbf{Ego-Exo4D} & \textbf{\textsc{MotorSense}} \\
  \cmidrule(lr){3-5} \cmidrule(lr){6-6} \cmidrule(lr){7-7} \cmidrule(lr){8-8}
   & \textbf{Pre-training} & \textbf{Verb} & \textbf{Noun} & \textbf{Action} & \textbf{Action} & \textbf{Keystep} & \textbf{Action} \\
  \midrule
  \multirow{6}{*}{\rotatebox{90}{\textit{Frozen}}}
   & None (K400-MAE init.)        & 43.85 & 17.74 & 9.93  & 29.17 & 17.72 & 16.86 \\
   & $\rho$SimCLR                 & 43.99 & 24.73 & 13.79 & 24.42 & 27.82 & 20.31 \\
   & $\rho$MoCo                   & 42.84 & 23.67 & 13.02 & 22.87 & 26.89 & 18.83 \\
   & $\rho$BYOL                   & 38.50 & 15.64 & 6.58  & 20.21 & 18.05 & 8.68  \\
  \cmidrule(lr){2-8}              
   & Ours            & \textbf{59.85} & \textbf{29.79} & \textbf{21.29} & \textbf{33.57} & \textbf{36.11} & \textbf{35.02} \\
  \cmidrule(lr){2-8}              
   & $\Delta$ vs.\ best baseline  & $+15.86$ & $+5.06$ & $+7.50$ & $+4.40$ & $+8.29$ & $+14.71$ \\
  \midrule
  \multirow{6}{*}{\rotatebox{90}{\textit{End-to-end}}}
   & None (K400-MAE init.)        & 70.42 & 50.01 & 40.06 & 67.57           & 45.82 & 42.22 \\
   & $\rho$SimCLR                 & 70.41 & 51.99 & 41.58 & 65.82$^\dagger$ & 49.48 & 42.83 \\
   & $\rho$MoCo                   & 70.35 & 52.16 & 41.85 & 65.36$^\dagger$ & 48.72 & 42.71 \\
   & $\rho$BYOL                   & 70.12 & 50.48 & 40.83 & 65.15$^\dagger$ & 46.13 & 39.51 \\
  \cmidrule(lr){2-8}              
   & Ours             & \textbf{74.09} & \textbf{52.25} & \textbf{43.81} & \textbf{68.76} & \textbf{51.47} & \textbf{47.51} \\
  \cmidrule(lr){2-8}              
   & $\Delta$ vs.\ best baseline  & $+3.67$ & $+0.09$ & $+1.96$ & $+1.19$ & $+1.99$ & $+4.68$ \\
  \bottomrule
  \end{tabular}%
  }
\caption{\textbf{Action recognition results across benchmarks.} Top-1 accuracy (\%) with the backbone frozen (top) or fine-tuned end-to-end (bottom) on EPIC-Kitchens-100~\citep{damen2018scaling}, Something-Something~v2~\citep{goyal2017something}, Ego-Exo4D~\citep{grauman2024egoexo4d}, and an annotated subset of \textsc{MotorSense}. All rows use a Hiera-B backbone initialized from K400-MAE~\citep{ryali2023hiera}; $\rho$SimCLR, $\rho$MoCo, and $\rho$BYOL~\citep{feichtenhofer2021large} are pre-trained on the same \textsc{MotorSense} recordings with RGB-only objectives. Best result per column in \textbf{bold}; $\Delta$ rows give our margin over the best baseline.}
\label{tab:action_recognition}
\end{table*}

\subsection{Cross-Modal Pre-training for Downstream Tasks}
\label{sec:downstream_eval_exp}

We pre-train video and image networks via cross-modal (video--EMG) contrastive alignment. For full pre-training implementation details, see Section \ref{appendix:contrastive_pretraining} within the Appendix. To evaluate whether the aligned backbone captures representations useful for physical action understanding, we fine-tune it over related downstream tasks. The projection heads are discarded and replaced with task-specific layers (a classification head for action recognition, and regression heads for 3D hand reconstruction and contact reconstruction).

\textbf{Action Recognition.} Table~\ref{tab:action_recognition} reports top-1 accuracy of different pre-training methods on action recognition. We fine-tune the contrastively pre-trained backbone on four downstream datasets chosen for their emphasis on hand--object manipulation: EPIC-Kitchens-100~\citep{damen2018scaling}, Something-Something~v2~\citep{goyal2017something}, Ego-Exo4D~\citep{grauman2024egoexo4d}, and a verb--noun annotated subset of \textsc{MotorSense}. For EPIC-Kitchens-100 we report verb, noun, and action accuracy; for the remaining datasets we report a single action-level metric. The projection heads of Section~\ref{sec:emg_pretraining} are discarded and replaced with a linear classification head, which is trained either with the backbone frozen or end-to-end. All models use a Hiera-B backbone initialized from the MAE-K400 weights released with Hiera~\citep{ryali2023hiera}. We compare against these weights directly, and against $\rho$SimCLR, $\rho$MoCo, and $\rho$BYOL~\citep{feichtenhofer2021large} pre-trained on the same \textsc{MotorSense} recordings and hand crops with RGB-only objectives, so that the comparison isolates the contribution of the EMG signal from that of additional in-domain video. Sensorimotor alignment outperforms every baseline on every benchmark, with the largest gains in the frozen setting. For ablations, see Section~\ref{sec:imu_ablation} in the Appendix.
 
\begin{table}[t]
\centering
\setlength{\tabcolsep}{5pt}
\renewcommand{\arraystretch}{1.1}
\small
\begin{tabular}{l cc cc cc}
\toprule
 & \multicolumn{2}{c}{\textbf{ARCTIC}} & \multicolumn{2}{c}{\textbf{H2O}} & \multicolumn{2}{c}{\textbf{HOT3D}} \\
\cmidrule(lr){2-3} \cmidrule(lr){4-5} \cmidrule(lr){6-7}
\textbf{Pre-training} & MPJPE $\downarrow$ & A-MPJPE $\downarrow$ & MPJPE $\downarrow$ & A-MPJPE $\downarrow$ & MPJPE $\downarrow$ & A-MPJPE $\downarrow$ \\
\midrule
Default                             & 24.29$^\dagger$ & 38.77$^\dagger$ & 24.28          & 51.20          & 30.85$^\dagger$          & 54.58$^\dagger$ \\
\textsc{MotorSense} (ours) & \textbf{21.71}  & \textbf{35.97}  & \textbf{19.65} & \textbf{44.04} & \textbf{28.39}$^\dagger$ & \textbf{48.85}$^\dagger$ \\
\midrule
$\Delta$                             & $-2.58$ & $-2.80$ & $-4.63$ & $-7.16$ & $-2.46$ & $-5.73$ \\
\bottomrule
\end{tabular}

\vspace{4pt}
\caption{\textbf{3D hand reconstruction.} Mean per-joint position error (mm; lower is better) relative to the hand root (MPJPE) and in camera coordinates without root alignment (A-MPJPE) on ARCTIC~\citep{fan2023arctic},
H2O~\citep{kwon2021h2o}, and HOT3D~\citep{banerjee2025hot3d}, using a hand-only
ArcticNet-SF~\citep{fan2023arctic} with a ResNet-50 backbone initialized from ImageNet alone or
after sensorimotor alignment on \textsc{MotorSense}. $\Delta$ is the change in error from our
pre-training.}
\label{tab:hand_localization}
\end{table}

\textbf{3D Hand Reconstruction.} We assess whether cross-modal pre-training improves recovery of fine-grained 3D hand geometry from egocentric video. Table~\ref{tab:hand_localization} reports results on
  three egocentric benchmarks: the head-mounted view of ARCTIC~\citep{fan2023arctic}, H2O~\citep{kwon2021h2o}, and HOT3D~\citep{banerjee2025hot3d}. We train a hand-only variant of ArcticNet-SF~\citep{fan2023arctic} (object branch removed) separately on each dataset with a ResNet-50 backbone, and compare
  initializing the backbone from ImageNet against our \textsc{MotorSense} pre-trained ResNet-50. All runs share the same architecture and training schedule; only the backbone initialization differs.
 
  \textbf{Hand Contact Reconstruction.} We next test whether cross-modal pre-training improves prediction of where the hand is in contact with its environment, information that EMG encodes directly
  through the forces exerted during manipulation. Table~\ref{tab:contact} reports results on the contact-annotated subset of \textsc{MotorSense} (Section~\ref{appendix:dataset_collection}), where the task
  is to predict, from a crop around one hand, which vertices are in contact, or that the hand is not in contact at all. We compare against training from ImageNet
  initialization and from scratch, and three image-level self-supervised methods learned over \textsc{MotorSense} videos, PixPro~\citep{xie2021propagate}, DenseCL~\citep{wang2021dense}, and
  PeCLR~\citep{spurr2021self}. All methods use a ResNet-50 backbone; we report per-vertex IoU and contact-detection AP, averaged over three seeds.

\subsection{Analysis of the Sensorimotor Embedding Space}
\label{sec:exp_sensorimotor_analysis}

\begin{wrapfigure}{r}{0.5\textwidth}
  \centering
  \includegraphics[width=0.5\textwidth]{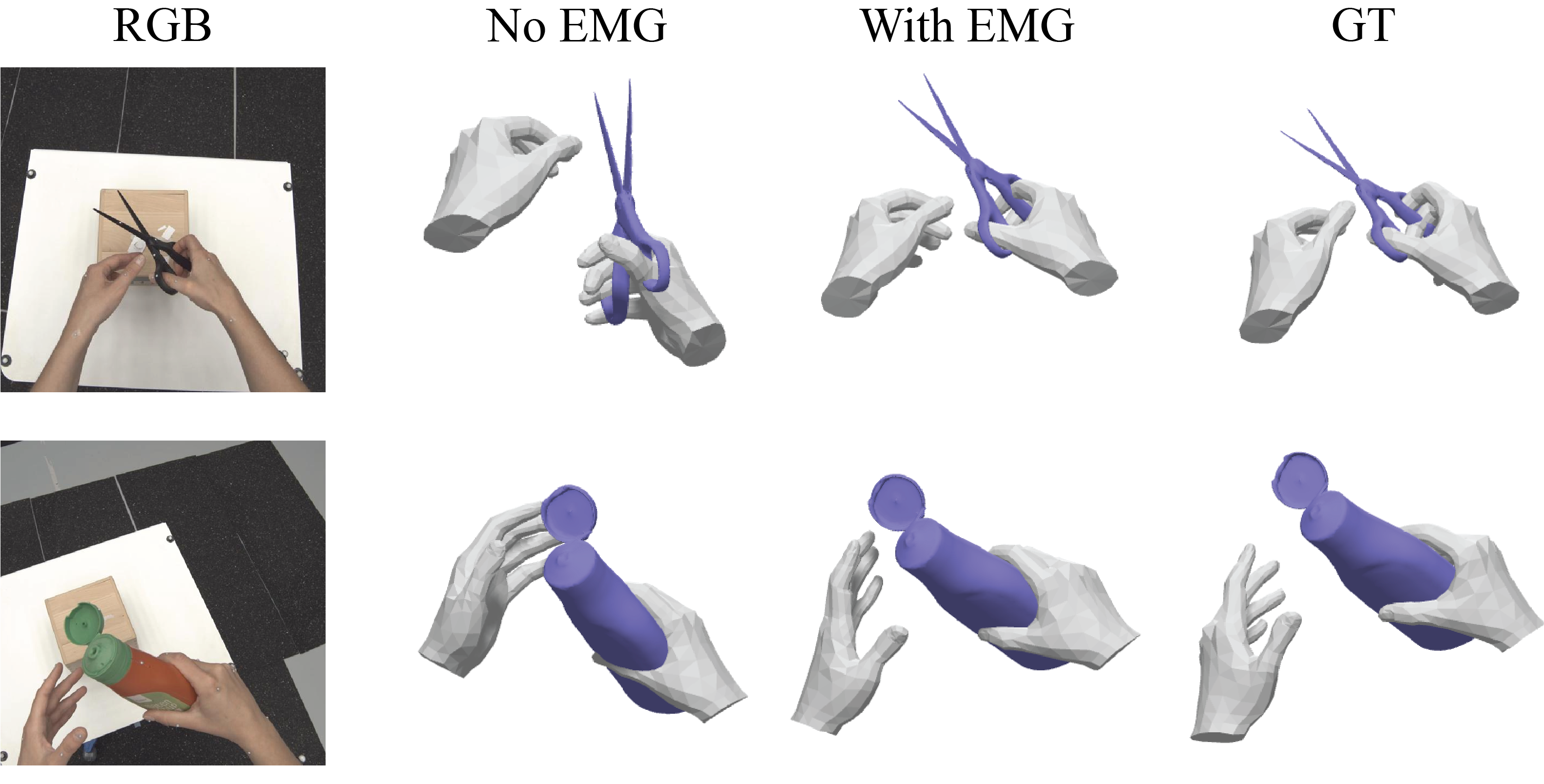}
  \caption{Qualitative results on ARCTIC \cite{fan2023arctic} for 3D hand-object reconstruction. Each row shows the RGB input, reconstructions from a vision backbone trained without EMG and with EMG (\textsc{MotorSense} pre-training), and ground truth. Incorporating EMG during pre-training yields 3D hands and objects that more closely match the ground truth.}
  \label{fig:3D_Hand_Object}
\end{wrapfigure}

We next examine whether sensorimotor alignment (Section~\ref{sec:emg_pretraining}) organizes video and EMG by hand action, with no action labels used during pre-training. We sample annotated windows from the \textsc{MotorSense} recordings, keeping those that lie within a single annotated action and contain a visible hand. We then encode the video clip and the bimanual EMG of each window with the two towers of the final pre-trained model. We embed the $L_2$-normalized video and EMG embeddings of these windows together with t-SNE~\citep{van2008visualizing}. For panel~(b), we pair video queries with their most similar EMG embedding by cosine similarity. The verb--noun annotations are used only to colour the projection; they play no role in pre-training or in computing the embeddings.

Figure~\ref{fig:sensorimotor_tsne} shows the result. Video and EMG form two separate clusters, the ``modality gap'' typical of contrastively aligned encoders~\citep{liang2022mind}, yet cross-modal nearest neighbours still match by hand action: each video query in panel~(b) retrieves EMG from a different recording that performs the same action, despite differences in object, scene, and participant. Within the EMG cluster (panel~(c), where a subset of actions is coloured by verb, with marker shape denoting the object), windows of the same action form compact regions, and actions that share a verb lie near one another (e.g.\ the \textit{cut} family), even though the model was never exposed to these labels.

\begin{wraptable}{r}{0.58\textwidth}
\centering
\vspace{-\baselineskip}
\setlength{\tabcolsep}{5pt}
\renewcommand{\arraystretch}{1.1}
\small
\begin{tabular}{lcc}
\toprule
\textbf{Pre-training} & \textbf{IoU (\%)} $\uparrow$ & \textbf{AUROC (\%)} $\uparrow$ \\
\midrule
ImageNet~\citep{deng2009imagenet} & 48.41 $\pm$ 0.76 & 92.08 $\pm$ 0.17 \\
DenseCL~\citep{wang2021dense}     & 48.56 $\pm$ 0.91 & 91.27 $\pm$ 0.29 \\
PeCLR~\citep{spurr2021self}       & 48.30 $\pm$ 0.56 & 91.71 $\pm$ 0.52 \\
PixPro~\citep{xie2021propagate}   & 50.59 $\pm$ 0.66 & 93.36 $\pm$ 0.25 \\
\midrule
Ours                 & \textbf{55.20} $\pm$ 0.62 & \textbf{97.19} $\pm$ 0.15 \\
\midrule
$\Delta$ vs.\ best baseline       & $+4.61$ & $+3.83$ \\
\bottomrule
\end{tabular}
\caption{\textbf{Hand contact reconstruction on \textsc{MotorSense}.} IoU (\%) between predicted and annotated contact regions on hands in contact, and AUROC (\%) for detecting whether a hand is in contact; mean $\pm$ std over three seeds. Best in \textbf{bold}; $\Delta$ is our margin over the best baseline.}
\label{tab:contact}
\end{wraptable}

\subsection{Motor Codes as a Control Signal for Video Diffusion}
\label{sec:exp_video_diffusion}

Table \ref{tab:video_diffusion} illustrates the results of different training and conditioning methods for video diffusion. Our method conditions on the discrete motor codes through a HyperNetwork-predicted LoRA. We compare against four baselines. \emph{Zero-shot I2V} and \emph{supervised I2V} receive only the reference frame, with the supervised variant additionally fine-tuned on \textsc{MotorSense} videos; together they ablate the contribution of motor-code conditioning. The \emph{text-prompt} baseline procedurally converts the raw EMG/IMU streams into a natural-language description that is fed as conditioning to the diffusion model, evaluating an alternative representation of the same underlying signal. The \emph{video-history} baseline replaces motor codes with a history of frames as input to the HyperNetwork-predicted LoRA, isolating the value of motor codes against a purely visual context. To assess visual fidelity we report Fréchet Video Distance (FVD)~\citep{unterthiner2019fvd}, along with per-frame LPIPS~\citep{zhang2018unreasonable} and PSNR. All metrics are computed on a bounding box centered around the hands rather than the full image, since the manipulation region is where conditioning differences are most visible. For qualitative visualizations of generated videos as well as full implementation details, see Section \ref{sec:appendix_video_diffusion} within the Appendix.

\begin{table}[t]
\centering
\setlength{\tabcolsep}{8pt}
\renewcommand{\arraystretch}{1.05}
\small
\begin{tabular}{lcccc}
\toprule
Conditioning & Trainable Params & FVD $\downarrow$ & LPIPS $\downarrow$ & PSNR $\uparrow$ \\
\midrule
CogVideoX-5B-I2V (supervised) & 5.57B & 772 & 0.35 & 17.3 \\
CogVideoX-5B-I2V (zero-shot) & 0 & 926 & 0.36 & 17.4 \\
\hspace{1em}$+$ language conditioning & 0 & 935 & 0.41 & 18.0 \\
\hspace{1em}\hspace{1.5em}$+$ video history & 250M & 373 & 0.30 & 22.0 \\
\midrule
Ours & 50M & \textbf{304} & \textbf{0.27} & \textbf{22.5} \\
\bottomrule
\end{tabular}
\vspace{8pt}
\caption{\textbf{Image-to-video generation on held-out \textsc{MotorSense} recordings}. All methods other than the top row share the same frozen video diffusion backbone and differ only in their conditioning signal and the trainable HyperNetwork/LoRA parameters. Evaluation is performed over a bounding box produced by Aria MPS \cite{engel2023project} (applied jointly to generated and original images). Best results are shown in \textbf{bold}.}
\label{tab:video_diffusion}
\end{table}

\section{Discussion}

\textbf{Sensorimotor alignment directly encodes action structure.}
The t-SNE projection in Figure~\ref{fig:sensorimotor_tsne} reveals that sensorimotor alignment organizes both video and EMG by action, despite being learned without manual labels. Windows of actions such as \textit{mix salad} and \textit{scrape bowl} form compact, cleanly separated regions, and actions sharing a verb but differing in object (e.g., \textit{cut}) sit close together in the embedding space. The downstream results further demonstrate this: the largest action-recognition gains appear under frozen evaluation ($+15.9$ verb accuracy on EPIC-Kitchens-100, $+14.7$ on \textsc{MotorSense}).

% \textbf{Structure of embeddings learned through sensorimotor alignment.} The t-SNE projection in Figure~\ref{fig:sensorimotor_tsne} reveals that sensorimotor alignment organizes both video and EMG by action, despite being learned without manual labels. Windows of actions such as \textit{mix salad} and \textit{scrape bowl} form compact, cleanly separated regions, and actions sharing a verb but differing in object (e.g., \textit{cut onion} and \textit{cut pepper}) sit close together in the embedding space. This action structure helps explain why the aligned video representations transfer well to action-centric downstream tasks.

\textbf{The gains come from EMG, rather than additional in-domain video.}
Matched RGB-only baselines trained on the same recordings isolate the contribution of EMG supervision from that of additional in-domain video. For action recognition, $\rho$SimCLR, $\rho$MoCo, and
$\rho$BYOL yield inconsistent gains and often fall below the K400-MAE initialization
(e.g., on SSv2), whereas our sensorimotor alignment learning outperforms every baseline on all four
benchmarks (Table~\ref{tab:action_recognition}). The same pattern holds at the image
level: DenseCL~\cite{wang2021dense}, PeCLR~\cite{spurr2021self}, and
PixPro~\cite{xie2021propagate} trained on \textsc{MotorSense} frames offer at most
marginal gains over ImageNet on contact reconstruction, while our model improves IoU by
$+4.6$ and AUROC by $+3.8$ over the best of them (Table~\ref{tab:contact}).

% \textbf{Sensorimotor alignment directly encodes action structure.}
% The largest action-recognition gains appear under frozen evaluation ($+15.9$ verb
% accuracy on EPIC-Kitchens-100, $+14.7$ on \textsc{MotorSense}), where the backbone
% cannot adapt to the downstream labels. That the gap narrows but persists under end-to-end
% fine-tuning suggests alignment with EMG instills a transferable prior rather than a
% feature the task head could recover on its own, reducing the labeled data needed
% downstream~\cite{moltisanti2017trespassing}. The embedding analysis in
% Figure~\ref{fig:sensorimotor_tsne} is consistent with this: windows of the same action form
% compact regions and verb families cluster, with no labels used in pre-training.

\textbf{EMG captures hand articulation and contact information.}
Because EMG measures finger articulation and exerted force while being robust to visual occlusion, we
expect its largest benefit on tasks where hands are partially occluded by the objects they
manipulate. Pre-training lowers joint error on all three 3D hand reconstruction benchmarks,
most notably on H2O ($-4.6$\,mm MPJPE, $-7.2$\,mm A-MPJPE;
Table~\ref{tab:hand_localization}, Figure~\ref{fig:3D_Hand_Object}), and yields the
contact-reconstruction improvements above, consistent with EMG directly encoding the
forces exerted at contact.

\textbf{Motor codes are an effective control signal for generation.}
In image-to-video generation (Table~\ref{tab:video_diffusion}), the zero-shot backbone
performs poorly and language conditioning adds little, confirming that text
under-specifies fine-grained manipulation. Both Video2LoRA-based
variants~\cite{wu2026video2lora} outperform full supervised fine-tuning, and motor-code
conditioning yields the highest fidelity for hand--object interaction, with the
qualitative differences (correct stirring vs.\ scooping, spreading vs.\ rotating,
finger placement on the avocado) shown in Figures~\ref{fig:diff_example_1}--\ref{fig:diff_example_3}.

\textbf{Limitations.} Our setup uses an 8-channel consumer-grade EMG band; while sufficient to demonstrate the value of aligning vision with EMG, higher-density bands (16 channels in commercial devices, upwards of 128 in research systems \cite{ting2021sensing}) would provide a richer signal as such hardware becomes more accessible. 

\textbf{Broader Impact.} As wearable EMG matures into a mainstream input modality, associated datasets and methods become increasingly impactful---supporting applications from VR/AR and personalized CGI to robot understanding of human collaborators, to high-stakes domains such as surgical training where virtual practice can ease reliance on scarce resources like cadavers. More broadly, motor-grounded representations open a path toward modeling hand action across diverse bodies and motor abilities, with implications for assistive and rehabilitative technologies.

\newpage
\bibliographystyle{splncs04}
\bibliography{main}

\newpage
\section{Appendix}

\subsection{Data Collection Protocol}
\label{sec:data_collection_protocol}

\begin{figure}[b!]
  \centering
  \includegraphics[width=\linewidth]{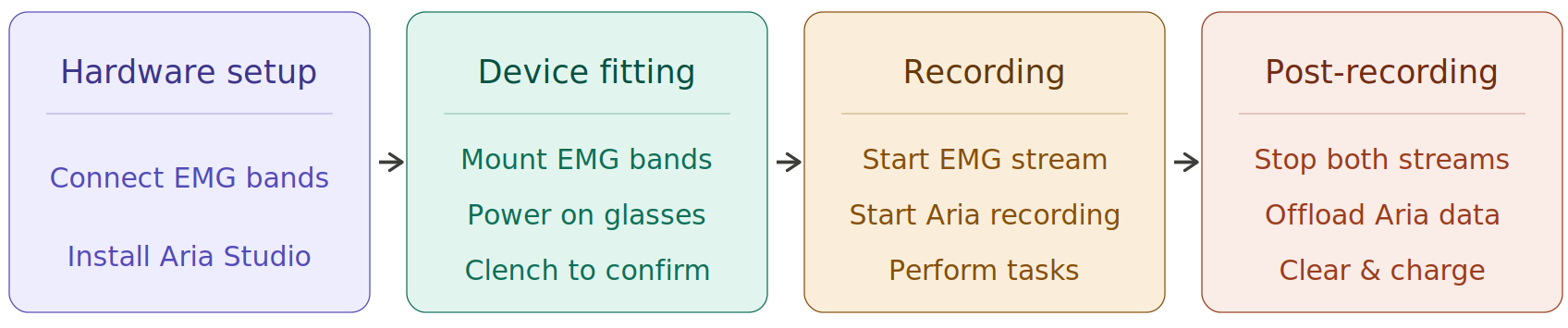}
  \caption{\textbf{The \textsc{MotorSense} data collection workflow.} Each session is organized into four stages, flowing left to right: \textit{hardware setup} (connecting the two MindRove armbands over separate Wi-Fi access points and verifying installation of Aria Studio), \textit{device fitting} (placing the bands slightly above each wrist, powering on and configuring the Aria glasses, and confirming correct mounting via a clench test), \textit{recording} (launching the EMG and Aria streams together and proceeding with unscripted manipulation), and \textit{post-recording} (stopping both streams, offloading Aria recordings via Aria Studio, and resetting the hardware for the next session).}
  \label{fig:data_collection_flow}
\end{figure}

Each \textsc{MotorSense} recording session takes approximately one hour from setup to teardown and is organized into four stages, summarized in Figure~\ref{fig:data_collection_flow}. In the \textbf{hardware setup} stage, the participant installs the EMG streaming code on a recording laptop and installs Meta's Aria Studio on the same machine for later transfer of glasses recordings. Because the two MindRove armbands stream over separate Wi-Fi access points, each band is connected from a dedicated laptop to avoid stream collision. In the \textbf{device fitting} stage, the bands are placed slightly above each wrist rather than along the forearm; participants with larger forearms are prioritized, as the bands slip on slender wrists and produce motion artifacts that render the data unusable. The Aria glasses are then powered on, paired over Bluetooth to the participant's phone, and configured with our custom profile (RGB at $15$ fps and $1440 \times 1440$, SLAM at $15$ fps, audio and GPS on, eye tracking off, internet and Bluetooth disabled during recording). Before any task begins, the participant performs an isolated clench with each hand to confirm that both bands are mounted correctly, verifying distinct activation spikes on the per-hand visualizers. In the \textbf{recording} stage, the EMG stream and the Aria recording (started from the phone app) are launched together, and the participant proceeds with unscripted manipulation tasks. In the \textbf{post-recording} stage, the EMG window is closed, the Aria recording is stopped from the phone, the recording is transferred to the laptop via Aria Studio and then deleted from the glasses, and the glasses are placed on charge for the next session.

 \subsection{Dataset Details}
  \label{appendix:dataset_collection}

  \textbf{Statistics.} \textsc{MotorSense} comprises $100$ hours of synchronized egocentric video and bimanual EMG + IMU, all centered on unscripted kitchen and household manipulation. Recordings come
  from three capture setups: Project Aria Gen~1 glasses ($18$ recordings, $22.0$ h), Project Aria Gen~2 glasses ($19$ recordings, $10.0$ h), and a head-mounted smartphone with an ultra-wide camera ($68.0$
  h). The Aria recordings were contributed by $8$ participants; the smartphone sessions are not attributed to participants. All recordings are used, without labels, for contrastive pre-training. Action
  segments are annotated on $85$ recordings ($5{,}209$ segments, $57$ verbs, $180$ nouns) and hand contact on $57$ smartphone recordings ($4{,}700$ intervals), with held-out validation recordings for
  each.
 
  \textbf{Capture Hardware.} Each forearm wears a MindRove armband with $8$ dry EMG electrodes and a $6$-axis IMU (accelerometer and gyroscope), streaming at $500$ Hz over Wi-Fi. For the egocentric visual
  stream we use Meta's Project Aria device \cite{engel2023project} (Gen~1 and Gen~2), whose forward-facing RGB camera is rectified to a pinhole model and resized ($704\times704$ for Gen~1;
  $1280\times960$ or $1008\times756$ for Gen~2), or a head-mounted smartphone recording $896\times504$ frames at $30$ fps. Video and EMG are placed on a single clock, so every frame is matched to the
  muscle activity of both arms.
 
  \textbf{Data Processing.} For Aria recordings, we leverage Meta's Machine Perception Services (MPS) to extract 6-DoF camera trajectories and 3D hand tracking, from which we derive per-frame hand
  bounding boxes; for smartphone recordings, hand boxes come from an off-the-shelf hand detector. Because the armbands stream over Wi-Fi, samples are occasionally lost; we drop training windows whose EMG
  contains gaps, saturation, or artifacts, removing about $4\%$ of candidate windows.
 
  To remove the low-frequency drift components from the EMG waveform, we apply a $4$th-order zero-phase Butterworth high-pass filter at $10$ Hz in line with existing approaches \cite{salter2024emg2pose}.
  We then z-score normalize each EMG and IMU channel over the recording. We release the raw EMG and IMU streams, and this preprocessing is applied by our data loaders.

\subsection{Implementation Details}

\subsubsection{Motor Code Discovery} 
\label{sec:appendix_code_discovery}

The \emph{temporal encoder} is a stack of strided one-dimensional convolutions interleaved with residual blocks, which downsample the signal along the time axis by a factor determined by a code rate of $64$ codes per second. Lower rates force each token to encode longer spans of motor activity, whereas higher rates force each token to encode finer-grained motor activity. A $1\times 1$ convolution then projects each encoder output into the codebook's embedding space, and the \emph{quantizer} replaces each projected latent with its nearest codebook entry; the codebook is updated by exponential moving averages of cluster assignments rather than by gradient descent, and gradients reach the encoder through a straight-through estimator (see \cite{van2017neural} for details). Finally, the \emph{WaveNet-style decoder} \cite{oord2016wavenet} upsamples the quantized latent sequence and reconstructs the input window.

We train the model with a reconstruction loss and a latent commitment penalty,
\begin{equation}
    \mathcal{L} = \lVert \hat{x} - x \rVert_2^2 + \lambda \, \lVert \operatorname{sg}[z] - q \rVert_2^2,
\end{equation}
where $x$ is the input window, $\hat{x}$ its reconstruction, $z$ the continuous pre-quantization latent, $q$ its vector-quantized counterpart (associated codebook embedding), $\operatorname{sg}[\cdot]$ the stop-gradient operator, and $\lambda$ the latent commitment weight; the codebook vectors themselves are maintained by the EMA updates above rather than by this loss.

The EMG data (sampled at $500$ Hz) is first preprocessed following Section ~\ref{appendix:dataset_collection} with a moving average window of $5$ samples. We then train separate motor-code models for the left and right hand on the EMG recordings of \textsc{MotorSense}. The VQ-VAE model ingests the processed EMG data over windows of length $1.5$ s. We use an embedding dimension $d = 256$, a codebook of $N = 64$ entries, and a target code rate (the number of codes per second) of $64$. The latent commitment weight is set to $\lambda = 0.25$. The model is optimized with a batch size of $128$ over a single RTX A5000 GPU.

\subsubsection{Video Diffusion}
\label{sec:appendix_video_diffusion}
We instantiate the conditioning pipeline of Section~\ref{sec:diffusion_methods} on top of the CogVideoX-Fun image-to-video backbone~\citep{yang2024cogvideox}, which we keep entirely frozen throughout training together with its T5 text encoder and CogVideoX VAE. Each training sample is a $49$-frame clip sampled at a frame rate of $15$ FPS (the native frame rate) from the training split \textsc{MotorSense} recordings, together with the paired bimanual motor-code sequences $\mathbf{k}_L, \mathbf{k}_R$ produced by the frozen tokenizers of Section~\ref{sec:diffusion_methods}. The clip's first frame is supplied as the image-to-video reference. The motor-code embedding network, the HyperNetwork, and the auxiliary LiLoRA matrices are the only trainable parameters. We train with AdamW at learning rate $1 \times 10^{-5}$, batch size $1$, for $30K$ steps on \textsc{MotorSense}.  All training was done over $4$ H200 GPUs, with 128 GB of RAM.\textbf{}

\subsubsection{Cross-Modal Contrastive Pre-training}
  \label{appendix:contrastive_pretraining}
  Pre-training operates on pairs of egocentric video and bimanual EMG + IMU drawn from all \textsc{MotorSense} recordings, with no action labels. Each sample is a $1.5$ s window (during training the
  window duration is sampled from $[1, 4.5]$ s in steps of $0.25$ s): the RGB video is of shape $16\times224\times224$, together with the bimanual EMG + IMU spanning the same interval extended by $1$ s of
  context on each side. Windows whose EMG fails our signal-quality checks are discarded. When a hand is visible, we apply a consistent temporal crop that contains the hand with a random scale $s \in
  (0.1, 1.0)$ of the frame (otherwise a random crop with $s \in (0.4, 1.0)$), followed by color jitter and RandAugment.
 
  The video tower is Hiera-Base initialized from MAE K400 pre-trained weights. We keep its $8$ temporal positions, average over space, and apply a linear projection to the shared embedding space of 
  dimension $1024$. EMG and IMU are encoded by two separate signal towers of identical design, operating on the high-pass filtered ($10$ Hz) and standardized $500$ Hz signals of both arms ($16$ EMG and
  $12$ IMU channels). Each tower applies a temporal convolution followed by two residual depthwise-separable temporal blocks that downsample to ${\sim}50$ Hz, a two-layer bidirectional LSTM over the full
  context window, and reads out the sequence at the $8$ video time steps, followed by LayerNorm and a linear projection to the same $1024$-d space. We apply no temporal pooling: each embedding is the
  flattened $8 \times 1024$ token sequence treated as a single vector. All outputs are $L_2$-normalized as single vectors, and their cosine similarities are divided by a fixed temperature $\tau = 0.01$.
  We optimize the average of two symmetric InfoNCE losses, video--EMG and video--IMU, with embeddings gathered across all devices. Each global batch is drawn from a single recording using non-overlapping
  windows, so that negatives share the scene and participant.
  
  We train with AdamW (weight decay $0.01$) under a cosine schedule with $3$ warmup epochs, using per-group learning rates (video backbone $5\times10^{-5}$, video projection $3\times10^{-4}$, EMG and IMU
  towers $3\times10^{-4}$), for $180$ epochs at a global batch size of $328$. Due to computational constraints, rather than jointly pre-training on K400 and \textsc{MotorSense}, we instead initialize from
  K400 pre-trained weights, align over \textsc{MotorSense}'s data, and transfer the resulting EMG-aware representations to downstream tasks. Training was done on $4$ H200 GPUs with $512$ GB of RAM.

  \subsection{Pre-training Ablations}                                                                                          
  \label{sec:ablations}
  \label{sec:imu_ablation}

  We ablate four design choices of our cross-modal pre-training: (i) the sensing modalities used for alignment, (ii) the number of EMG electrodes, (iii) how batches are formed, and (iv) whether embeddings
  are pooled over time before the contrastive loss. We transfer each backbone to action recognition on EPIC-KITCHENS-100 and MotorSense under frozen and
  fine-tuned evaluation, following Section~\ref{sec:downstream_eval_exp}, with a single seed per evaluation.               

   \begin{table}[h]
  \centering
  \setlength{\tabcolsep}{6pt}
  \renewcommand{\arraystretch}{1.1}
  \small
  \begin{tabular}{l l c c c c}
  \hline
  & & \multicolumn{2}{c}{\textbf{EPIC-KITCHENS-100}} & \multicolumn{2}{c}{\textbf{MotorSense}} \\
  \cmidrule(lr){3-4} \cmidrule(lr){5-6}                                                      
  \textbf{Ablation} & \textbf{Variant} & \textbf{Frozen} & \textbf{Fine-tuned} & \textbf{Frozen} & \textbf{Fine-tuned} \\
  \hline
  --          & K400 init.          & 9.93           & 40.06          & 16.86          & 42.22 \\
  --          & Ours (reference)    & \textbf{21.29}          & \textbf{43.81}          & \textbf{35.02}          & \textbf{47.51} \\
  \hline
  Modalities  & EMG only            & 16.33          & 41.27          & 30.46          & 45.29 \\
              & IMU only            & 15.36          & 41.10          & 25.48          & 43.26 \\
  \hline
  Electrodes  & 4 per arm           & 18.88          & 41.86          & 31.57          & 44.62 \\
              & 2 per arm           & 19.29 & 41.80          & 31.82          & 44.98 \\
  \hline
  Batching    & Mixed recordings    & 18.21          & 42.20 & 29.29          & 46.46 \\
  \hline
  Pooling     & Temporal mean       & 17.64          & 41.84          & 32.55 & 44.86 \\
  \hline
  \end{tabular}                                                                        
  \vspace{8pt}

  \caption{\textbf{Pre-training ablations.} Action top-1 accuracy on EPIC-KITCHENS-100 and MotorSense with a Hiera-B backbone; each ablated row changes a single setting of the reference (EMG+IMU, 8 electrodes per arm, same-recording batches, time-resolved
  embeddings). Single seed per evaluation. Best pre-trained result per column in \textbf{bold}.}
  \label{tab:ablations}                                                           
  \end{table}

\subsection{Video Diffusion Sequences}
\label{sec:video_diffusion_seq}

\begin{figure*}[b!]
  \centering
  \includegraphics[width=1\textwidth]{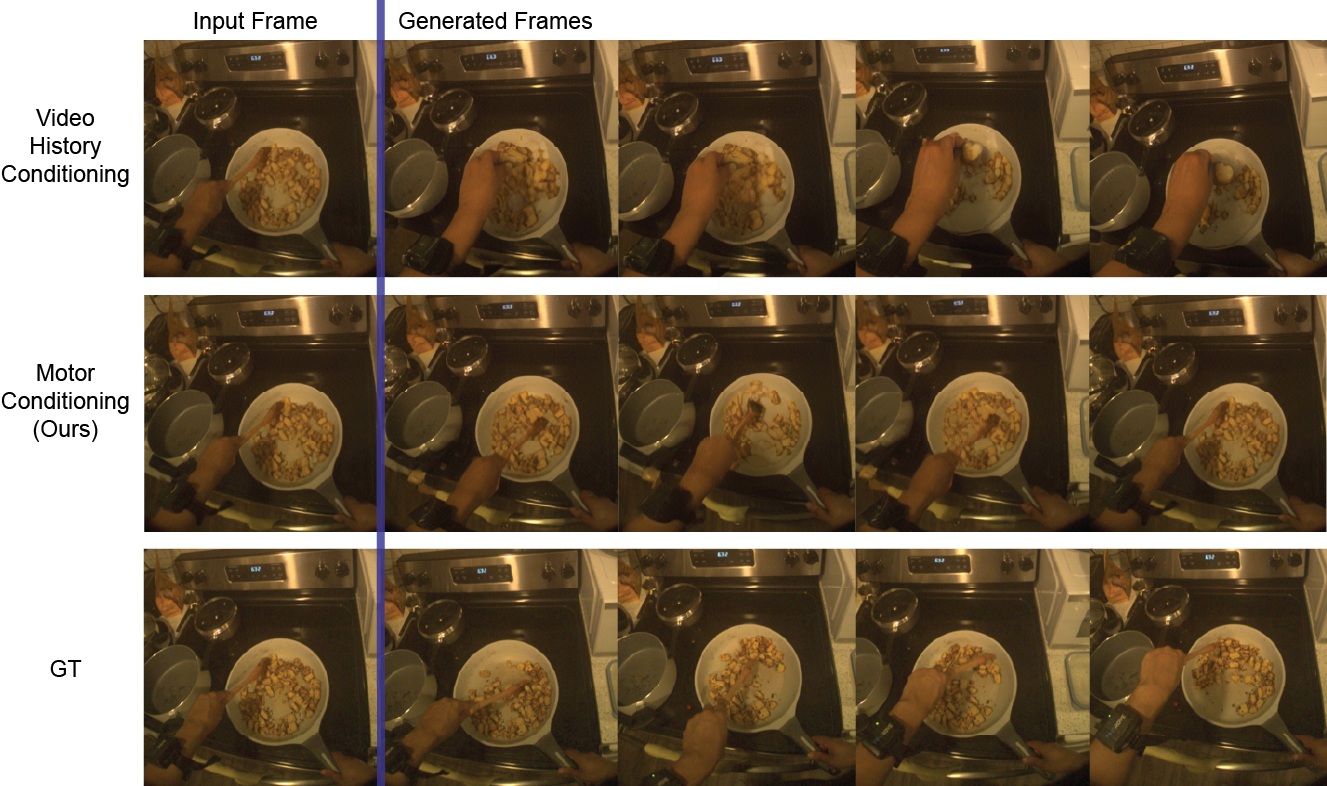}
  \caption{\textbf{Qualitative results for video diffusion}. We compare our proposed motor conditioning to a video diffusion model conditioned solely on the video history (unimodal conditioning), for a sequence involving the stirring of tofu in a pan. The video conditioning model misinterprets the action performed, generating a video of a scooping action as opposed to a stirring action, whereas our method correctly generates a video of the stirring of tofu.}
  \label{fig:diff_example_1}
\end{figure*}

\begin{figure*}[t!]
  \centering
  \includegraphics[width=1\textwidth]{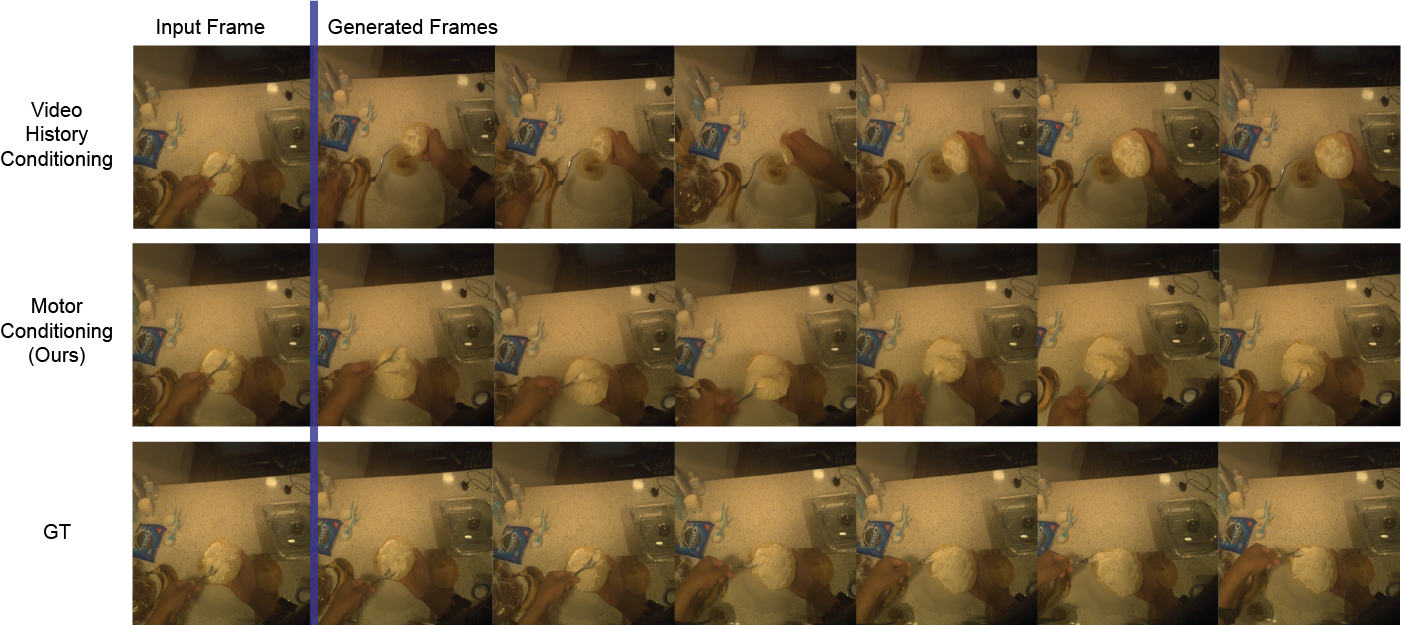}
  \caption{\textbf{Qualitative results for video diffusion}. We compare our proposed motor conditioning to a video diffusion model conditioned solely on the video history (unimodal conditioning), for a sequence involving the spreading of mayonnaise with a spoon. The video conditioning model misinterprets the action performed, generating a rotation of the bread as opposed to a spreading action of the spoon, whereas our method correctly generates a video of the spreading of mayonnaise with a spoon.}
  
  \label{fig:diff_example_2}
\end{figure*}

\begin{figure*}[t!]
  \centering
  \includegraphics[width=1\textwidth]{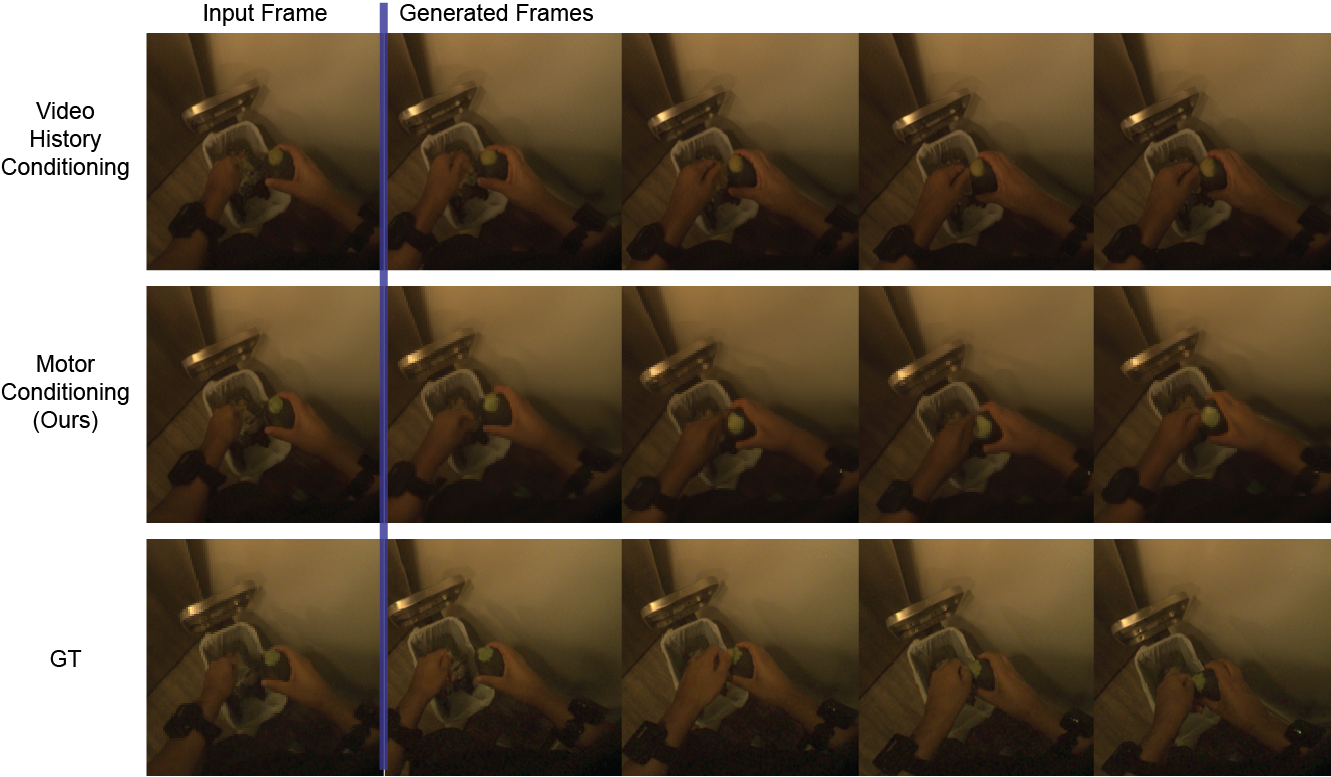}
  \caption{\textbf{Qualitative results for video diffusion}. We compare our proposed motor conditioning to a video diffusion model conditioned solely on the video history (unimodal conditioning), for a bimanual manipulation sequence involving the peeling of an avocado. The video conditioning model renders the index finger as resting against the thumb, whereas our method correctly positions the index finger against the surface of the avocado for the peeling action.}
  \label{fig:diff_example_3}
\end{figure*}

In Figures  \ref{fig:diff_example_1}, \ref{fig:diff_example_2},  and \ref{fig:diff_example_3} we show qualitative visualizations comparing the original RGB video with the generated video over both the \textit{Video History Conditioning} as well as our proposed \textit{Motor Code Conditioning}. Visualizations and evaluations are performed over bounding box crops centered on the hands, as opposed to the entirety of the original RGB video.

\clearpage
\section*{NeurIPS Paper Checklist}

\begin{enumerate}

\item {\bf Claims}
    \item[] Question: Do the main claims made in the abstract and introduction accurately reflect the paper's contributions and scope?
    \item[] Answer: \answerYes{} % Replace by \answerYes{}, \answerNo{}, or \answerNA{}.
    \item[] Justification: The paper introduces \textsc{MotorSense}, a dataset pairing egocentric video with bimanual surface EMG and IMU, and evaluates representations learned from it across action recognition, 3D hand reconstruction, hand contact reconstruction, and motor-conditioned video diffusion. Following the review process, the dataset claims have been narrowed: we no longer describe \textsc{MotorSense} as the first paired video--EMG dataset (ActionSense~\cite{delpreto2022actionsense} precedes it; the differences are discussed in Section~\ref{sec:related_work} and the Appendix), we state explicitly that the participant pool is not population-representative, and claims about cross-subject generalization are limited to what the held-out-participant and cross-domain evaluations in the Appendix support. All experimental claims are covered in Section~\ref{sec:experiments}, with implementation details and ablations in the Appendix.
    \item[] Guidelines:
    \begin{itemize}
        \item The answer \answerNA{} means that the abstract and introduction do not include the claims made in the paper.
        \item The abstract and/or introduction should clearly state the claims made, including the contributions made in the paper and important assumptions and limitations. A \answerNo{} or \answerNA{} answer to this question will not be perceived well by the reviewers. 
        \item The claims made should match theoretical and experimental results, and reflect how much the results can be expected to generalize to other settings. 
        \item It is fine to include aspirational goals as motivation as long as it is clear that these goals are not attained by the paper. 
    \end{itemize}

\item {\bf Limitations}
    \item[] Question: Does the paper discuss the limitations of the work performed by the authors?
    \item[] Answer: \answerYes{} % Replace by \answerYes{}, \answerNo{}, or \answerNA{}.
    \item[] Justification: Limitations are discussed in the Discussion section and in an expanded Limitations section in the Appendix. Beyond the 8-channel consumer-grade EMG hardware, we discuss: (i) the small participant pool; (ii) selection bias with respect to forearm circumference, which arises from the mechanical mounting of the MindRove bands (the bands slip on slender wrists and produce unusable data) and which correlates with sex and body size, so that \textsc{MotorSense} should not be read as population-representative; (iii) the effect of anatomical variation on EMG amplitude and electrode--muscle correspondence, and its implications for motor-code prediction; (iv) the scope of our cross-subject evaluation, which covers motor-code prediction on held-out participants but not downstream subject-disjoint evaluation of action recognition or 3D reconstruction; (v) the larger performance drop under activity-domain shift (assembly, cleaning, laundry) than under subject shift; (vi) the role of IMU in stabilizing the EMG signal under rapid wrist motion; and (vii) privacy and biometric considerations of releasing egocentric video and physiological EMG. Dataset statistics and compute are reported in the Experiments section and Appendix.
    \item[] Guidelines:
    \begin{itemize}
        \item The answer \answerNA{} means that the paper has no limitation while the answer \answerNo{} means that the paper has limitations, but those are not discussed in the paper. 
        \item The authors are encouraged to create a separate ``Limitations'' section in their paper.
        \item The paper should point out any strong assumptions and how robust the results are to violations of these assumptions (e.g., independence assumptions, noiseless settings, model well-specification, asymptotic approximations only holding locally). The authors should reflect on how these assumptions might be violated in practice and what the implications would be.
        \item The authors should reflect on the scope of the claims made, e.g., if the approach was only tested on a few datasets or with a few runs. In general, empirical results often depend on implicit assumptions, which should be articulated.
        \item The authors should reflect on the factors that influence the performance of the approach. For example, a facial recognition algorithm may perform poorly when image resolution is low or images are taken in low lighting. Or a speech-to-text system might not be used reliably to provide closed captions for online lectures because it fails to handle technical jargon.
        \item The authors should discuss the computational efficiency of the proposed algorithms and how they scale with dataset size.
        \item If applicable, the authors should discuss possible limitations of their approach to address problems of privacy and fairness.
        \item While the authors might fear that complete honesty about limitations might be used by reviewers as grounds for rejection, a worse outcome might be that reviewers discover limitations that aren't acknowledged in the paper. The authors should use their best judgment and recognize that individual actions in favor of transparency play an important role in developing norms that preserve the integrity of the community. Reviewers will be specifically instructed to not penalize honesty concerning limitations.
    \end{itemize}

\item {\bf Theory assumptions and proofs}
    \item[] Question: For each theoretical result, does the paper provide the full set of assumptions and a complete (and correct) proof?
    \item[] Answer: \answerNA{} % Replace by \answerYes{}, \answerNo{}, or \answerNA{}.
    \item[] Justification: No theoretical results were included. Our paper is not on the theory track.
    \item[] Guidelines:
    \begin{itemize}
        \item The answer \answerNA{} means that the paper does not include theoretical results. 
        \item All the theorems, formulas, and proofs in the paper should be numbered and cross-referenced.
        \item All assumptions should be clearly stated or referenced in the statement of any theorems.
        \item The proofs can either appear in the main paper or the supplemental material, but if they appear in the supplemental material, the authors are encouraged to provide a short proof sketch to provide intuition. 
        \item Inversely, any informal proof provided in the core of the paper should be complemented by formal proofs provided in appendix or supplemental material.
        \item Theorems and Lemmas that the proof relies upon should be properly referenced. 
    \end{itemize}

    \item {\bf Experimental result reproducibility}
    \item[] Question: Does the paper fully disclose all the information needed to reproduce the main experimental results of the paper to the extent that it affects the main claims and/or conclusions of the paper (regardless of whether the code and data are provided or not)?
    \item[] Answer: \answerYes{} % Replace by \answerYes{}, \answerNo{}, or \answerNA{}.
    \item[] Justification: The Methods, Experiments, and Appendix contain all information needed to reproduce our results, including data preprocessing, architectures, optimizers, schedules, and splits. The dataset, trained models, and training code are released (see Questions 5 and 13). For implementation details inherited from prior work, we refer the reader to the relevant source throughout the paper (e.g., "we refer the reader to Video2LoRA \cite{wu2026video2lora} for...").
    \item[] Guidelines:
    \begin{itemize}
        \item The answer \answerNA{} means that the paper does not include experiments.
        \item If the paper includes experiments, a \answerNo{} answer to this question will not be perceived well by the reviewers: Making the paper reproducible is important, regardless of whether the code and data are provided or not.
        \item If the contribution is a dataset and\slash or model, the authors should describe the steps taken to make their results reproducible or verifiable. 
        \item Depending on the contribution, reproducibility can be accomplished in various ways. For example, if the contribution is a novel architecture, describing the architecture fully might suffice, or if the contribution is a specific model and empirical evaluation, it may be necessary to either make it possible for others to replicate the model with the same dataset, or provide access to the model. In general. releasing code and data is often one good way to accomplish this, but reproducibility can also be provided via detailed instructions for how to replicate the results, access to a hosted model (e.g., in the case of a large language model), releasing of a model checkpoint, or other means that are appropriate to the research performed.
        \item While NeurIPS does not require releasing code, the conference does require all submissions to provide some reasonable avenue for reproducibility, which may depend on the nature of the contribution. For example
        \begin{enumerate}
            \item If the contribution is primarily a new algorithm, the paper should make it clear how to reproduce that algorithm.
            \item If the contribution is primarily a new model architecture, the paper should describe the architecture clearly and fully.
            \item If the contribution is a new model (e.g., a large language model), then there should either be a way to access this model for reproducing the results or a way to reproduce the model (e.g., with an open-source dataset or instructions for how to construct the dataset).
            \item We recognize that reproducibility may be tricky in some cases, in which case authors are welcome to describe the particular way they provide for reproducibility. In the case of closed-source models, it may be that access to the model is limited in some way (e.g., to registered users), but it should be possible for other researchers to have some path to reproducing or verifying the results.
        \end{enumerate}
    \end{itemize}

\item {\bf Open access to data and code}
    \item[] Question: Does the paper provide open access to the data and code, with sufficient instructions to faithfully reproduce the main experimental results, as described in supplemental material?
    \item[] Answer: \answerYes{} % Replace by \answerYes{}, \answerNo{}, or \answerNA{}.
    \item[] Justification: We release the \textsc{MotorSense} dataset (synchronized video, bimanual EMG and IMU, audio, derived spectrograms, and MPS-derived camera and hand pose), the trained models, and the training and evaluation code, together with instructions for reproducing the experiments, at \url{https://anonymous.4open.science/r/MotorSense-7173/}. % TODO: replace with the de-anonymized URL for the camera-ready version.
    \item[] Guidelines:
    \begin{itemize}
        \item The answer \answerNA{} means that paper does not include experiments requiring code.
        \item Please see the NeurIPS code and data submission guidelines (\url{https://neurips.cc/public/guides/CodeSubmissionPolicy}) for more details.
        \item While we encourage the release of code and data, we understand that this might not be possible, so \answerNo{} is an acceptable answer. Papers cannot be rejected simply for not including code, unless this is central to the contribution (e.g., for a new open-source benchmark).
        \item The instructions should contain the exact command and environment needed to run to reproduce the results. See the NeurIPS code and data submission guidelines (\url{https://neurips.cc/public/guides/CodeSubmissionPolicy}) for more details.
        \item The authors should provide instructions on data access and preparation, including how to access the raw data, preprocessed data, intermediate data, and generated data, etc.
        \item The authors should provide scripts to reproduce all experimental results for the new proposed method and baselines. If only a subset of experiments are reproducible, they should state which ones are omitted from the script and why.
        \item At submission time, to preserve anonymity, the authors should release anonymized versions (if applicable).
        \item Providing as much information as possible in supplemental material (appended to the paper) is recommended, but including URLs to data and code is permitted.
    \end{itemize}

\item {\bf Experimental setting/details}
    \item[] Question: Does the paper specify all the training and test details (e.g., data splits, hyperparameters, how they were chosen, type of optimizer) necessary to understand the results?
    \item[] Answer: \answerYes{} % Replace by \answerYes{}, \answerNo{}, or \answerNA{}.
    \item[] Justification: Data splits (including the held-out-environment validation split and the held-out-participant and cross-domain evaluation splits), hyperparameters, optimizers, schedules, and model selection criteria are given in the Experiments section and in the Implementation Details and ablation sections of the Appendix.
    \item[] Guidelines:
    \begin{itemize}
        \item The answer \answerNA{} means that the paper does not include experiments.
        \item The experimental setting should be presented in the core of the paper to a level of detail that is necessary to appreciate the results and make sense of them.
        \item The full details can be provided either with the code, in appendix, or as supplemental material.
    \end{itemize}

\item {\bf Experiment statistical significance}
    \item[] Question: Does the paper report error bars suitably and correctly defined or other appropriate information about the statistical significance of the experiments?
    \item[] Answer: \answerYes{} % Replace by \answerYes{}, \answerNo{}, or \answerNA{}.
    \item[] Justification: Error bars (mean $\pm$ standard deviation over three training seeds) are reported for the hand contact reconstruction experiments (Table~\ref{tab:contact}).
    \item[] Guidelines:
    \begin{itemize}
        \item The answer \answerNA{} means that the paper does not include experiments.
        \item The authors should answer \answerYes{} if the results are accompanied by error bars, confidence intervals, or statistical significance tests, at least for the experiments that support the main claims of the paper.
        \item The factors of variability that the error bars are capturing should be clearly stated (for example, train/test split, initialization, random drawing of some parameter, or overall run with given experimental conditions).
        \item The method for calculating the error bars should be explained (closed form formula, call to a library function, bootstrap, etc.)
        \item The assumptions made should be given (e.g., Normally distributed errors).
        \item It should be clear whether the error bar is the standard deviation or the standard error of the mean.
        \item It is OK to report 1-sigma error bars, but one should state it. The authors should preferably report a 2-sigma error bar than state that they have a 96\% CI, if the hypothesis of Normality of errors is not verified.
        \item For asymmetric distributions, the authors should be careful not to show in tables or figures symmetric error bars that would yield results that are out of range (e.g., negative error rates).
        \item If error bars are reported in tables or plots, the authors should explain in the text how they were calculated and reference the corresponding figures or tables in the text.
    \end{itemize}

\item {\bf Experiments compute resources}
    \item[] Question: For each experiment, does the paper provide sufficient information on the computer resources (type of compute workers, memory, time of execution) needed to reproduce the experiments?
    \item[] Answer: \answerYes{} % Replace by \answerYes{}, \answerNo{}, or \answerNA{}.
    \item[] Justification: The GPU type, number of GPUs, and memory used for each experiment (motor-code discovery, cross-modal pre-training, and video diffusion) are reported in the Implementation Details section of the Appendix.
    \item[] Guidelines:
    \begin{itemize}
        \item The answer \answerNA{} means that the paper does not include experiments.
        \item The paper should indicate the type of compute workers CPU or GPU, internal cluster, or cloud provider, including relevant memory and storage.
        \item The paper should provide the amount of compute required for each of the individual experimental runs as well as estimate the total compute. 
        \item The paper should disclose whether the full research project required more compute than the experiments reported in the paper (e.g., preliminary or failed experiments that didn't make it into the paper). 
    \end{itemize}
    
\item {\bf Code of ethics}
    \item[] Question: Does the research conducted in the paper conform, in every respect, with the NeurIPS Code of Ethics \url{https://neurips.cc/public/EthicsGuidelines}?
    \item[] Answer: \answerYes{} % Replace by \answerYes{}, \answerNo{}, or \answerNA{}.
    \item[] Justification: The data collection was reviewed and approved by our institution's IRB; all participants gave written informed consent covering research use, public release, and reproduction of frames and clips in publications; participants were compensated; risks were disclosed; and the released data were screened and de-identified before release. These procedures are documented in the Appendix and in Questions 11, 14, and 15 below.
    \item[] Guidelines:
    \begin{itemize}
        \item The answer \answerNA{} means that the authors have not reviewed the NeurIPS Code of Ethics.
        \item If the authors answer \answerNo, they should explain the special circumstances that require a deviation from the Code of Ethics.
        \item The authors should make sure to preserve anonymity (e.g., if there is a special consideration due to laws or regulations in their jurisdiction).
    \end{itemize}

\item {\bf Broader impacts}
    \item[] Question: Does the paper discuss both potential positive societal impacts and negative societal impacts of the work performed?
    \item[] Answer: \answerYes{} % Replace by \answerYes{}, \answerNo{}, or \answerNA{}.
    \item[] Justification: Positive impacts (VR/AR, assistive and rehabilitative technology, surgical training, human--robot collaboration) are discussed in the Broader Impact paragraph of the Discussion. Negative impacts are discussed in the Limitations section of the Appendix: the privacy risk of egocentric recording in home environments, the sensitivity of EMG as a physiological signal, and the risk that models trained on a non-representative participant pool perform worse for under-represented body types. The mitigations we applied to the release are described under Safeguards below.
    \item[] Guidelines:
    \begin{itemize}
        \item The answer \answerNA{} means that there is no societal impact of the work performed.
        \item If the authors answer \answerNA{} or \answerNo, they should explain why their work has no societal impact or why the paper does not address societal impact.
        \item Examples of negative societal impacts include potential malicious or unintended uses (e.g., disinformation, generating fake profiles, surveillance), fairness considerations (e.g., deployment of technologies that could make decisions that unfairly impact specific groups), privacy considerations, and security considerations.
        \item The conference expects that many papers will be foundational research and not tied to particular applications, let alone deployments. However, if there is a direct path to any negative applications, the authors should point it out. For example, it is legitimate to point out that an improvement in the quality of generative models could be used to generate Deepfakes for disinformation. On the other hand, it is not needed to point out that a generic algorithm for optimizing neural networks could enable people to train models that generate Deepfakes faster.
        \item The authors should consider possible harms that could arise when the technology is being used as intended and functioning correctly, harms that could arise when the technology is being used as intended but gives incorrect results, and harms following from (intentional or unintentional) misuse of the technology.
        \item If there are negative societal impacts, the authors could also discuss possible mitigation strategies (e.g., gated release of models, providing defenses in addition to attacks, mechanisms for monitoring misuse, mechanisms to monitor how a system learns from feedback over time, improving the efficiency and accessibility of ML).
    \end{itemize}
    
\item {\bf Safeguards}
    \item[] Question: Does the paper describe safeguards that have been put in place for responsible release of data or models that have a high risk for misuse (e.g., pre-trained language models, image generators, or scraped datasets)?
    \item[] Answer: \answerYes{} % Replace by \answerYes{}, \answerNo{}, or \answerNA{}.
    \item[] Justification: The released dataset contains egocentric video recorded in participants' homes, so we applied a documented release protocol described in the Appendix: participants recorded alone and were instructed that no other person be present, and manual review confirmed that no bystanders appear in the released recordings; participants were instructed not to speak, and the small number of segments containing inadvertent speech were removed; location and raw GPS metadata are not included in the release; and every recording was screened and de-identified before release by a combination of manual review and EgoBlur~\cite{raina2023egoblur}, targeting faces and other identifying content. The dataset is released under research-use terms stated in the Appendix and alongside the data.
    \item[] Guidelines:
    \begin{itemize}
        \item The answer \answerNA{} means that the paper poses no such risks.
        \item Released models that have a high risk for misuse or dual-use should be released with necessary safeguards to allow for controlled use of the model, for example by requiring that users adhere to usage guidelines or restrictions to access the model or implementing safety filters. 
        \item Datasets that have been scraped from the Internet could pose safety risks. The authors should describe how they avoided releasing unsafe images.
        \item We recognize that providing effective safeguards is challenging, and many papers do not require this, but we encourage authors to take this into account and make a best faith effort.
    \end{itemize}

\item {\bf Licenses for existing assets}
    \item[] Question: Are the creators or original owners of assets (e.g., code, data, models), used in the paper, properly credited and are the license and terms of use explicitly mentioned and properly respected?
    \item[] Answer: \answerYes{} % Replace by \answerYes{}, \answerNo{}, or \answerNA{}.
    \item[] Justification: All existing datasets (EPIC-Kitchens-100~\cite{damen2018scaling}, Something-Something v2~\cite{goyal2017something}, Ego-Exo4D~\cite{grauman2024egoexo4d}, Meccano~\cite{ragusa2021meccano}, ARCTIC~\cite{fan2023arctic}, H2O~\cite{kwon2021h2o}, HOT3D~\cite{banerjee2025hot3d}), pre-trained models (Hiera~\cite{ryali2023hiera}, EgoVideo, DINOv3, CogVideoX~\cite{yang2024cogvideox}), code (Video2LoRA~\cite{wu2026video2lora}, and hardware/services (Project Aria and MPS~\cite{engel2023project}, MindRove) are cited where used. % TODO: add the per-asset license table to the Appendix before camera-ready.
    \item[] Guidelines:
    \begin{itemize}
        \item The answer \answerNA{} means that the paper does not use existing assets.
        \item The authors should cite the original paper that produced the code package or dataset.
        \item The authors should state which version of the asset is used and, if possible, include a URL.
        \item The name of the license (e.g., CC-BY 4.0) should be included for each asset.
        \item For scraped data from a particular source (e.g., website), the copyright and terms of service of that source should be provided.
        \item If assets are released, the license, copyright information, and terms of use in the package should be provided. For popular datasets, \url{paperswithcode.com/datasets} has curated licenses for some datasets. Their licensing guide can help determine the license of a dataset.
        \item For existing datasets that are re-packaged, both the original license and the license of the derived asset (if it has changed) should be provided.
        \item If this information is not available online, the authors are encouraged to reach out to the asset's creators.
    \end{itemize}

\item {\bf New assets}
    \item[] Question: Are new assets introduced in the paper well documented and is the documentation provided alongside the assets?
    \item[] Answer: \answerYes{} % Replace by \answerYes{}, \answerNo{}, or \answerNA{}.
    \item[] Justification: We release three new assets: the \textsc{MotorSense} dataset, the trained models (EMG motor-code tokenizers, video--EMG aligned backbones, and the motor-conditioned diffusion adapters), and the training and evaluation code. The dataset is documented in the Appendix (collection protocol, participant instructions, statistics, preprocessing, splits, consent and de-identification procedures, known limitations and selection bias, and license/terms of use), and the same documentation, together with usage instructions, is provided alongside the assets at the release URL. Consent for research use and public release was obtained in writing from every participant.
    \item[] Guidelines:
    \begin{itemize}
        \item The answer \answerNA{} means that the paper does not release new assets.
        \item Researchers should communicate the details of the dataset\slash code\slash model as part of their submissions via structured templates. This includes details about training, license, limitations, etc. 
        \item The paper should discuss whether and how consent was obtained from people whose asset is used.
        \item At submission time, remember to anonymize your assets (if applicable). You can either create an anonymized URL or include an anonymized zip file.
    \end{itemize}

\item {\bf Crowdsourcing and research with human subjects}
    \item[] Question: For crowdsourcing experiments and research with human subjects, does the paper include the full text of instructions given to participants and screenshots, if applicable, as well as details about compensation (if any)? 
    \item[] Answer: \answerYes{} % Replace by \answerYes{}, \answerNo{}, or \answerNA{}.
    \item[] Justification: The full instructions given to participants are included in the Appendix. They cover installation and connection of the EMG streaming software, band placement on each wrist, the clench verification test used to confirm correct mounting, Aria glasses configuration and recording start/stop, session length, the requirement to record alone with no other person present and without speaking, and the post-recording offload procedure. Participants were recruited by convenience through word of mouth within the authors' institutional network of students, screened only for device-fit suitability and willingness to record cooking activity, and received a fixed payment of \$25 per recording session plus reimbursement for food used during cooking.
    \item[] Guidelines:
    \begin{itemize}
        \item The answer \answerNA{} means that the paper does not involve crowdsourcing nor research with human subjects.
        \item Including this information in the supplemental material is fine, but if the main contribution of the paper involves human subjects, then as much detail as possible should be included in the main paper. 
        \item According to the NeurIPS Code of Ethics, workers involved in data collection, curation, or other labor should be paid at least the minimum wage in the country of the data collector. 
    \end{itemize}

\item {\bf Institutional review board (IRB) approvals or equivalent for research with human subjects}
    \item[] Question: Does the paper describe potential risks incurred by study participants, whether such risks were disclosed to the subjects, and whether Institutional Review Board (IRB) approvals (or an equivalent approval/review based on the requirements of your country or institution) were obtained?
    \item[] Answer: \answerYes{} % Replace by \answerYes{}, \answerNo{}, or \answerNA{}.
    \item[] Justification: The data collection was reviewed and approved by our institution's IRB under two sequential protocols: the first protocol expired during the project timeline, and a second protocol was approved before the corresponding later sessions began, so every stage of collection was covered by an active approval. Participants were informed of the risks of the procedure before recording: minor skin irritation from the dry electrodes and band pressure, discomfort from extended wear, and the privacy risk of egocentric recording in their own home environment. All participants gave written informed consent covering participation, research use of the recorded video, audio, EMG, IMU, and derived data, public release of that data as a research dataset, and reproduction of individual frames and clips in publications and presentations, and were told they could withdraw at any point during collection. Full details are in the Appendix.
    \item[] Guidelines:
    \begin{itemize}
        \item The answer \answerNA{} means that the paper does not involve crowdsourcing nor research with human subjects.
        \item Depending on the country in which research is conducted, IRB approval (or equivalent) may be required for any human subjects research. If you obtained IRB approval, you should clearly state this in the paper. 
        \item We recognize that the procedures for this may vary significantly between institutions and locations, and we expect authors to adhere to the NeurIPS Code of Ethics and the guidelines for their institution. 
        \item For initial submissions, do not include any information that would break anonymity (if applicable), such as the institution conducting the review.
    \end{itemize}

\item {\bf Declaration of LLM usage}
    \item[] Question: Does the paper describe the usage of LLMs if it is an important, original, or non-standard component of the core methods in this research? Note that if the LLM is used only for writing, editing, or formatting purposes and does \emph{not} impact the core methodology, scientific rigor, or originality of the research, declaration is not required.
    %this research? 
    \item[] Answer: \answerNA{} % Replace by \answerYes{}, \answerNo{}, or \answerNA{}.
    \item[] Justification: LLMs were used only for data processing and filtering and for writing assistance; they are not a component of the proposed method, the motor-code representation, the contrastive pre-training, or the evaluation.
    \item[] Guidelines:
    \begin{itemize}
        \item The answer \answerNA{} means that the core method development in this research does not involve LLMs as any important, original, or non-standard components.
        \item Please refer to our LLM policy in the NeurIPS handbook for what should or should not be described.
    \end{itemize}

\end{enumerate}
\end{document}